\documentclass{article}

\usepackage[
]
{emnlp2021}

\usepackage{times}
\usepackage{latexsym}

\usepackage[utf8]{inputenc} % allow utf-8 input
\usepackage[T1]{fontenc}    % use 8-bit T1 fonts
\usepackage{tipa}           % get reasonable foreign name diacritics in bib
\usepackage{url}            % simple URL typesetting
\usepackage{booktabs}       % professional-quality tables
\usepackage{amsfonts}       % blackboard math symbols
\usepackage{nicefrac}       % compact symbols for 1/2, etc.
\usepackage{microtype}      % microtypography
\usepackage{xcolor}         % colors
\usepackage{multirow}
\usepackage{amsmath}
\usepackage{amssymb}
\usepackage{array} % to get fixed length right aligned columns for table 3
\usepackage{graphicx}
\graphicspath{{./images/}} 
\usepackage{enumitem}
\usepackage{bbm}
\usepackage[ruled, vlined, linesnumbered]{algorithm2e}
\usepackage[noend]{algpseudocode}
\newcommand{\D}{\mathcal{D}}

\newcommand{\Stilde}{\widetilde{\mathcal{S}}}
\newcommand{\set}[1]{\left\{#1\right\}}
\renewcommand{\SS}{\mathcal{S}}
\newcommand{\A}{\mathcal{A}}

\newcommand{\dataset}{\textbf{P}erturbation \textbf{A}ugmentation \textbf{N}LP \textbf{DA}taset} % name tbd
\newcommand{\acronym}{PANDA}
\newcommand{\perturber}{perturber} %berturber

\newcommand{\fairscore}{fairscore} % name tbd
\newcommand{\Fairscore}{Fairscore} % name tbd
\newcommand{\fairberta}{FairBERTa} % name tbd

\usepackage{tikz}
\usepackage{pgf,pgfplots,pgfplotstable}
\pgfplotsset{compat=1.17}
\usepackage{xcolor}

\definecolor{color1}{RGB}{213, 183, 94}
\definecolor{color2}{RGB}{255, 195, 0}
\definecolor{color3}{RGB}{255, 87, 51}
\definecolor{color4}{RGB}{144, 12, 63}
\definecolor{color5}{RGB}{213, 115,93}
\definecolor{color6}{RGB}{161, 21, 9}
\newcommand*\circled[1]{\tikz[baseline=(char.base)]{\node[shape=circle,draw,inner sep=1pt] (char) {#1};}}

\newcommand*{\yellowemph}[1]{%
  \tikz[baseline=(X.base)] \node[rectangle, fill=yellow, %rounded corners,
  inner sep=0.5mm] (X) {#1};%
}

\newcommand*{\blueemph}[1]{%
  \tikz[baseline=(X.base)] \node[rectangle, fill=blue!40 , %rounded corners,
  inner sep=0.5mm] (X) {#1};%
}

\usepackage{hyperref}       % hyperlinks
\hypersetup{
    colorlinks=true,
    allcolors=blue,
}

\usepackage{adjustbox}

\usepackage[noend]{algcompatible}

\newcommand{\emo}[1]{\raise1.0ex\hbox{{\normalfont \textrm{\scriptsize #1}}}} % Use: Author Name\emo{1, *}
\title{Perturbation Augmentation %with Demographic Perturbations 
for Fairer NLP}

\author{%
Rebecca Qian$^\dagger$ \quad Candace Ross$^\dagger$ \quad Jude Fernandes$^\dagger$  \quad \\
\bf Eric Smith$^\dagger$ \quad \bf Douwe Kiela\emo{$\ddagger$, *}\quad \bf Adina Williams\emo{$\dagger$, *} \\
  $^\dagger$ Facebook AI Research; $^\ddagger$ Hugging Face\\
  \texttt{rebeccaqian,adinawilliams@fb.com} \\
}

\begin{document}

\maketitle

\begin{abstract}
  Unwanted and often harmful social biases are becoming ever more salient in NLP research, affecting both models and datasets. In this work, we ask whether training on demographically perturbed data leads to fairer language models. We collect a large dataset of human annotated text perturbations and train a neural perturbation model, which we show outperforms heuristic alternatives. We find that (i) language models (LMs) pre-trained on demographically perturbed corpora are typically more fair, and (ii) LMs finetuned on perturbed GLUE datasets exhibit less demographic bias on downstream tasks, and (iii) fairness improvements do not come at the expense of performance on downstream tasks. Lastly, we discuss outstanding questions about how best to evaluate the (un)fairness of large language models. We hope that this exploration of neural demographic perturbation will help drive more improvement towards fairer NLP.
\end{abstract}

\section{Introduction}\label{sec:intro}

\begin{figure*}[h!] %placeholder fig. sketch
\centering
    \includegraphics[width=0.95\textwidth]{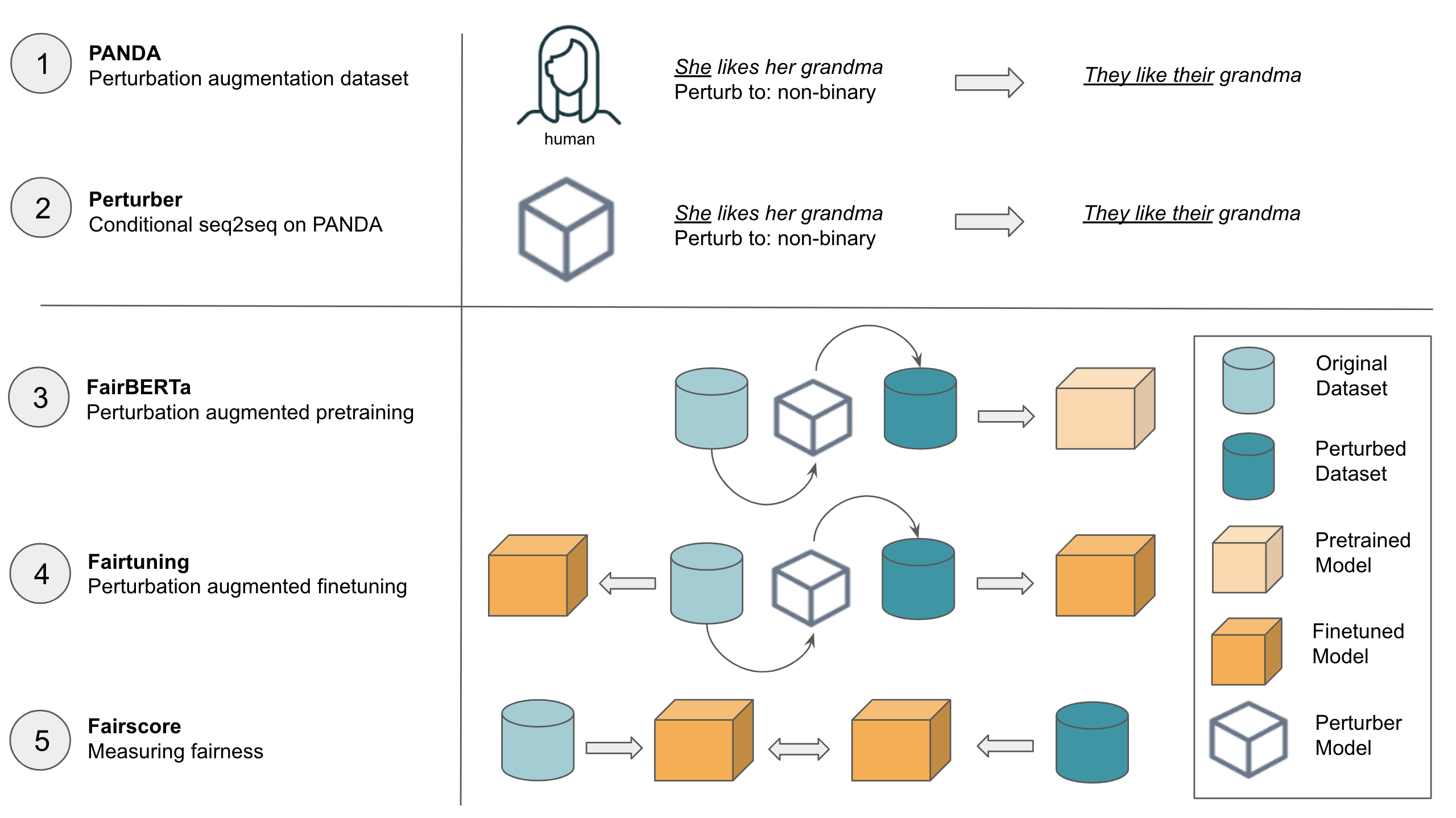}
    \caption{Our contributions.
    \circled{1} refers to our large scale annotated dataset (\acronym) of demographic perturbations. Our \perturber~in \circled{2} is trained on \acronym~to generate high quality perturbed text. In \circled{3}, we train a LM on data that has been augmented using the \perturber. In \circled{4}, we illustrate a method for finetuning on perturbation augmented validation data, which we call \emph{fairtuning}. Finally, we propose the \fairscore~\circled{5}, an extrinsic metric that quantifies fairness in LMs as robustness to demographic perturbation.
    }
    \label{fig:summary}
\end{figure*}

There is increasing evidence that models can instantiate social biases~\citep{buolamwini-gebru-2018-gender, stock-cisse-2018-convnets, fan2019plainsight,  merullo-etal-2019, prates-etal-2020-assessing}, often replicating or amplifying harmful statistical associations in their training data~\citep{caliskan-etal-2017-semantics, chang-etal-2019-bias}. Training models on data with representational issues can lead to unfair or poor treatment of particular demographic groups~\cite{barocas-etal-2017-problem, mehrabi-etal-2021-lawyers}, a problem that is particularly egregious for historically marginalized groups, including people of color~\cite{field-etal-2021-survey}, and women~\cite{hendricks-etal-2018-women}. As NLP moves towards training models on ever larger data samples~\citep{kaplan-etal-2020-scaling}, such data-related risks may grow~\citep{bender2021dangers}.% Even worse, models often fail to acknowledge the existence of some groups, such as non-binary individuals \cite{dev-etal-2021-harms}, altogether. Such exclusion and erasure is a clear harm that should be remedied.

In this work, we explore the efficacy of a dataset alteration technique that rewrites demographic references in text, such as changing ``women like shopping'' to ``men like shopping''. Similar demographic perturbation approaches have been fruitfully used to measure and often lessen the severity of social bias in text data~\citep{hall-maudslay-etal-2019-name, prabhakaran-etal-2019-perturbation,  zmigrod-etal-2019-counterfactual,  dinan-etal-2020-queens, dinan-etal-2020-multi, webster-etal-2020-measuring, ma-etal-2021-dynaboard, smith-etal-2021-hi, renduchintala-williams-2022-investigating, emmery-etal-2022-cyberbullying}. Most approaches for perturbing demographic references, however, rely on rule-based systems, which unfortunately tend to be rigid and error prone, resulting in noisy and unnatural perturbations (see \autoref{sec:heuristics}). While some have suggested that a neural demographic perturbation model may generate higher quality text rewrites, there are currently no annotated datasets large enough for training neural models \citep{sun-etal-2021-they}. 

In this work, we collect the first large-scale dataset of 98K human-generated demographic text perturbations, the \textbf{P}erturbation \textbf{A}ugmentation \textbf{N}LP \textbf{DA}taset~(\textbf{\acronym}). We use \acronym~to train a seq2seq controlled generation model, the \textbf{\perturber}. The \perturber~takes in (i) a source text snippet, (ii) a word in the snippet referring to a demographic group, and (iii) a new target demographic attribute, and generates a perturbed snippet that refers to the target demographic attribute, while preserving overall meaning. We find that the \perturber~generates high quality perturbations, outperforming heuristic alternatives. We use our neural \perturber~to augment existing training data with demographically altered examples, weakening unwanted demographic associations. 

% In past work, models that were trained on data where most or all shoppers are women learned the stereotype that ``women like shopping'' \cite{zhao-etal-2017-men}. 

We explore the effect of demographic perturbation on language model training both during \emph{pretraining} and \emph{finetuning} stages. We pretrain \textbf{\fairberta}, the first large language model trained on demographically perturbed corpora, and show that its fairness is improved, without degrading performance on downstream tasks. 

We also investigate the effect of \textbf{fairtuning}, i.e. finetuning models on perturbation augmented datasets, on model fairness. We find that fairtuned models perform well on a variety of natural language understanding~(NLU) tasks while also being fairer on average than models finetuned on the original, unperturbed datasets.

Finally, we propose \textbf{\fairscore}, an extrinsic fairness metric that uses the \perturber~to measure fairness as robustness to demographic perturbation. Given an NLU classification task, we define the \fairscore~as the change in model predictions between the original evaluation dataset and the perturbation augmented version. Prior approaches to measuring fairness in classifiers often rely on ``challenge datasets'' to measure how predictions differ in response to demographic changes in inputs~\citep{zhao-etal-2018-gender,rudinger2018gender,DeArteaga2019BiasIB,parrish-etal-2021-bbq}. However, collecting human annotations can be costly, and task specific evaluation sets do not always generalize across NLU tasks. The \fairscore~is a versatile, complementary method to challenge datasets that can be easily applied to any NLP dataset. We see significant improvements in the \fairscore~from fairtuning on a range of GLUE tasks.

% We find that fairtuned models show less demographic bias and perform well on a variety of standard NLU tasks. Our training results are promising, suggesting that perturbation augmentation can improve fairness in NLP datasets.

% Finally,

Our main contributions are summarized in \autoref{fig:summary}. Using a neural \perturber~to demographically augment model training data is a promising direction for lessening bias in large language models. To enable more exploration and improvement upon the present work, we will release \acronym, our controllable \perturber, \fairberta, and all other trained models and code artifacts under a permissive license.%\footnote{See our repository to download \acronym~and for more information: \href{https://github.com/facebookresearch/ResponsibleNLP}{github.com/facebookresearch/ResponsibleNLP}}

\section{Approach}\label{sec:approach}

We begin perturbation with a set of text snippets, each of which contains at least one \textbf{demographic term}. Demographic terms could be a pronoun (\textit{she, him, their}, etc.), a proper name (\textit{Sue, Yutong, Jamal}), a noun (\textit{son, grandparent}), an adjective labeling a demographic group (\textit{Asian, Black}) or another part of speech with demographic information. Each term instantiates one or more \textbf{demographic axes}, such as gender, each of which has several \textbf{demographic attributes}, such as ``man'', ``woman'', and ``non-binary/underspecified''. For each snippet, we perturb the demographic term to a new demographic attribute along its axis while preserving coreference information. If we consider the phrase ``\textit{women like shopping}'' where we select the demographic term ``\textit{women}'', we could perturb the sentence along the gender axis to refer to the gender attribute ``\textit{man}'', resulting in ``\textit{men like shopping}''. We use the following demographic axes and attributes: Gender (Man, Woman, Non-Binary/Underspecified), Race/Ethnicity\footnote{We use \href{https://www.census.gov/newsroom/blogs/random-samplings/2021/08/measuring-racial-ethnic-diversity-2020-census.html}{the US Census Survey} for race and ethnicity attributes; for a discussion of limitations arising from relying on the U.S. Census for race/ethnicity attributes, see \autoref{sec:limitations}.} (White, Black, Hispanic or Latino, Asian, Native American or Alaska Native, Hawaiian or Pacific Islander), and Age (Child < 18, Young 18-44, Middle-aged 45-64, Senior 65+, Adult Unspecified).

We avoid perturbing terms such as \textit{surgeon} or \textit{pink} that can be proxies for demographic axes (i.e., they have only statistical and/or stereotypical gender associations), precisely because our procedure aims to break statistical associations of this sort. While names are also only proxies for race and/or gender, we include them as demographic terms because names-based demographic associations have been shown to benefit from counterfactual augmentation~\citep{hall-maudslay-etal-2019-name, prabhakaran-etal-2019-perturbation, smith-etal-2021-hi}.

% If no demographic terms appear in a snippet, it will not be perturbed. 

One consequence of our approach is that not all factual content will be preserved through demographic perturbation (see \autoref{sec:limitations} for more discussion). %For example, we see the factual Queen Victoria perturbed to a counterfactual King Victor in the example, \textit{To whom did \textcolor{blue}{King Victor} lament that marriage was a shocking alternative to \textcolor{blue}{his} mother's presence? <SEP> Though \textcolor{blue}{king}, as an unmarried young \textcolor{blue}{man Victor} was required by social convention to live with \textcolor{blue}{his} mother, \textellipsis \textcolor{blue}{He} showed interest in Albert's education for the future role he would have to play as \textcolor{blue}{his} husband, but \textcolor{blue}{he} resisted attempts to rush \textcolor{blue}{him} into wedlock} from \acronym. 
In a research context, we are under no strict obligation to replicate in exact detail the world as it currently exists: for instance, we could create counterfactual text describing an alternative past where the first human on the moon was nonbinary. Our work is not a caveat-free endorsement of demographic perturbation, nor is it a blanket suggestion to apply it to all tasks in NLP. Nonetheless, we feel our research is relevant for answering the question: can demographically perturbed data be useful for improving the fairness of language models? We do not want to enable ``fairwashing'' by creating a simple but incomplete test that models can pass so as to be deemed ``safe''. Instead, the present work is something of an existence proof for the utility of neural demographic perturbation.

\paragraph{Formalizing Demographic Perturbation Augmentation:} Let $\SS$ be the input dataset consisting of variable-length snippets of text, 
where $s \in \SS$ is a text snippet and $w$ is a word in $s$ with demographic attribute $a_w$. Let $\A$ be a set of demographic attributes and $\mathcal{P} \subseteq \A \times \A $ be the set of \textit{(source, target)} attribute pairs where~$(a_s, a_t) \in \mathcal{P}$ defines one pair. We use $\mathcal{P}_\text{d}$ to denote the subset of attribute pairs that are under the demographic axis $d$, where $d \in \{gender,race,age\}$. For example, for $d = gender$, example attribute pairs for $\mathcal{P}_{gender}$ include \textit{(man, woman)}, \textit{(woman, non-binary)}. $\D_\text{d}$ denotes the dictionary of words for the demographic axis $d$. 

We illustrate the procedure for perturbation augmentation in \autoref{algo:DataCollection}. We sample text snippets to be used as inputs to the perturber from an existing text dataset $\mathcal{S}$. For each snippet $s \in \mathcal{S}$, we identify the set of perturbable demographic words using our words list. For each perturbable word $w$, we identify source and target demographic attributes for perturbation. For example, for $w = lady$, possible source and target attribute pairs include \textit{(woman, man)} and \textit{(woman, non-binary)}. We then sample a word and target attribute with uniform probability\footnote{For the finetuning datasets, we use a modified frequency-based sampling strategy that ensures representation of race/ethnicity perturbations, preserving dataset size.}, to preserve dataset size $|\mathcal{S}|$. 

\begin{algorithm}
\DontPrintSemicolon
 \textbf{Input:} dataset $\SS$, set of attribute pairs $\mathcal{P}_\text{d}$, dictionary of demographic words $\D_\text{d}$  \;
    
    \textbf{Initialize:} new dataset $\Stilde \leftarrow \emptyset$\;
    \For{snippet $s \in \SS$}{
        new snippet $\widetilde{s} \leftarrow s$\;
        new $K \leftarrow \emptyset$\;
        \For{word $w \in s \cap \D_\text{d}$}{
            \For{ $(\cdot,t) \in \set{(a_s, a_t) \in \mathcal{P}_{\text{d}} | a_s = a_w, a_s \neq a_t}$}{
            $K \leftarrow K \cup \set{w, t}$\;
            }
        }
        $(w, t) \sim \mathcal{U}(K)$\;
        $\widetilde{s} \leftarrow$ \texttt{perturber(s, w, t)}\;
        $\Stilde \leftarrow \Stilde \cup \set{\widetilde{s}}$\;
    }
    \textbf{Output:} $\Stilde$
  \caption{Data Augmentation via Demographic Perturbation}\label{algo:DataCollection}
\end{algorithm}

%\paragraph{Applying demographic perturbation to pretraining and finetuning:}

% {\color{color6}
% \paragraph{Applying perturbation augmentation to model training:}

\paragraph{Defining fairtuning:} Demographic perturbation augmentation is a flexible, scalable method that can be used to alter demographic representations in large training datasets. We explore the effects of demographic perturbation on model training in two settings: (i) pretraining large LMs on perturbation augmented datasets, and (ii) finetuning models on perturbation augmented NLU task datasets, an approach we refer to as \textbf{fairtuning}. In the supervised fairtuning setting, we apply the \perturber~to each training example, following~\autoref{algo:DataCollection}. For a labeled training dataset $D = \{x^{i}, y^{i}\}$ and perturber model $f_{P}$, we create a perturbation augmented dataset $\widetilde{D} = \{f_{P}(x^{i}), y^{i}\}$ that preserves the original label. We preserve the size of the dataset during perturbation augmentation to ensure fair comparisons.% with the original finetuning dataset.

% Given the scale and cost of pretraining, this may not be the most accessible approach for all scenarios. % We therefore also introduce \textbf{fairtuning}, an approach that uses the perturber to debias LMs during finetuning instead during of pretraining. Fairtuning is closely related to counterfactual generation \citep{wu-etal-2021-polyjuice}. 

% }

% We fairtune RoBERTa and \fairberta~ on the same 6 datasets in GLUE.

\paragraph{Defining the \fairscore:} We next define a fairness metric to measure robustness to demographic perturbation on classification tasks. Following~\newcite{prabhakaran-etal-2019-perturbation, ma-etal-2021-dynaboard, thrush-etal-2022-dynatask}, we assume that perturbing demographic references should have minimal to no effect on most of the NLU tasks we investigate. For instance, the sentiment of a review like \textit{Sue's restaurant was to die for} shouldn't be altered if we replace \textit{Sue} with \textit{Yitong}, as names shouldn't have any sentiment on their own, and the part of the text that does (i.e., \textit{...'s restaurant was to die for}) remains unchanged \citep{prabhakaran-etal-2019-perturbation}.
Models that utilize demographic terms as lexical ``shortcuts'' \citep{geirhos-etal-2020-shortcut} during classification will have a larger change in their predictions than models that do not, with the latter being deemed ``more fair'' by our metric. 

We measure how sensitive a model finetuned on a downstream classification task is to demographic perturbation by evaluating it on both the original evaluation set and a demographically altered version. The \textbf{\fairscore~}of a classifier is defined as the percentage of predictions that differ when the input is demographically altered.\footnote{We filter for examples containing demographic information in validation sets, to ensure that the fairscore is computed only on examples containing demographic information.} More formally, for a \perturber~model $f_{P}$ and text snippet $x$, let $\widetilde{x} \sim f_{P}(x)$ be the demographically altered perturber output. A classifier $f_{C}$ exhibits bias if for some input $x$ and demographically perturbed input $\widetilde{x}$, the predictions $f_{C}(x) \neq f_{C}(\widetilde{x})$. Given a classifier $f_{C}$ and an evaluation set $X$, we define the \fairscore~$F_{S}$ as
\setlength{\abovedisplayskip}{4.55pt}% hacky
\setlength{\belowdisplayskip}{4.55pt}
\begin{equation}
    F_{S}(f_{C}, X) = \frac{|\{x \in X|\ f_{C}(x) \neq f_{C}(\widetilde{x})\}|}{|X|}.
\end{equation}%
In the following sections, we describe how we collected an annotated dataset of text perturbations, trained a neural perturbation model, and show results of its applications to model training and bias measurement.

\section{\dataset~(\acronym)}\label{sec:dataset}
\begin{table*}[h!]
  \begin{center}
  \begin{adjustbox}{width=\linewidth}
  \begin{tabular}{p{8cm}p{1cm}p{2cm}p{8cm}p{8cm}}
    \toprule
    %\multicolumn{1}{c}{}                   \\
    \bf original source text & \bf selected word & \bf target (axis:attribute) & \bf human perturbed output (gold) & \bf \perturber~output \\ \toprule
    To whom did \textcolor{red}{Queen Victoria} lament that marriage was a shocking alternative to \textcolor{red}{\textbf{her}} mother's prescence? <SEP> Though \textcolor{red}{queen}, as an unmarried young \textcolor{red}{woman Victoria} was required by social convention to live with \textcolor{red}{her} mother, despite their differences over the Kensington System and \textcolor{red}{her} mother's continued reliance on Conroy. \textcolor{red}{Her} mother was consigned to a remote apartment in Buckingham Palace, and \textcolor{red}{Victoria} often refused to see \textcolor{red}{her}. When \textcolor{red}{Victoria} complained to Melbourne that \textcolor{red}{her} mother's close proximity promised ``torment for many years'', Melbourne sympathised but said it could be avoided by marriage, which \textcolor{red}{Victoria} called a ``schocking [sic] alternative''. \textcolor{red}{She} showed interest in Albert's education for the future role he would have to play as \textcolor{red}{her} husband, but \textcolor{red}{she} resisted attempts to rush \textcolor{red}{her} into wedlock.  & her &  gender:man &  To whom did \textcolor{blue}{King Victor} lament that marriage was a shocking alternative to \textcolor{blue}{his} mother's presence? <SEP> Though \textcolor{blue}{king}, as an unmarried young \textcolor{blue}{man Victor} was required by social convention to live with \textcolor{blue}{his} mother, despite their differences over the Kensington System and \textcolor{blue}{his} mother's continued reliance on Conroy. \textcolor{blue}{His} mother was consigned to a remote apartment in Buckingham Palace, and \textcolor{blue}{Victor} often refused to see her. When \textcolor{blue}{Victor} complained to Melbourne that \textcolor{blue}{his} mother's close proximity promised ``torment for many years'', Melbourne sympathized but said it could be avoided by marriage, which \textcolor{blue}{Victor} called a ``\textcolor{blue}{shocking} [ sic ] alternative''. \textcolor{blue}{He} showed interest in Albert's education for the future role he would have to play as \textcolor{blue}{his} husband, but \textcolor{blue}{he} resisted attempts to rush \textcolor{blue}{him} into wedlock. & To whom did \textcolor{blue}{King Victor} lament that marriage was a shocking alternative to \textcolor{blue}{his} mother's presence? <SEP> Though \textcolor{blue}{king}, as an unmarried young \textcolor{blue}{man Victor} was required by social convention to live with \textcolor{blue}{his} mother, despite their differences over the Kensington System and \textcolor{blue}{his} mother's continued reliance on Conroy. \textcolor{blue}{His} mother was consigned to a remote apartment in Buckingham Palace, and \textcolor{blue}{Victor} often refused to see her. When \textcolor{blue}{Victor} complained to Melbourne that \textcolor{blue}{his} mother's close proximity promised ``torment for many years'', Melbourne sympathized but said it could be avoided by marriage, which \textcolor{blue}{Victor} called a ``\textcolor{blue}{shocking} [ sic ] alternative''. \textcolor{blue}{He} showed interest in Albert's education for the future role he would have to play as \textcolor{blue}{his} husband, but \textcolor{blue}{he} resisted attempts to rush \textcolor{blue}{him} into wedlock. \\ 
    \midrule
    A ``\textcolor{red}{\textbf{black}} Austin Powers ?'' & black & race:asian & A\textcolor{blue}{n} ``\textcolor{blue}{Asian} Austin Powers?'' & A ``\textcolor{blue}{Asian} Austin Powers?'' \\
 \midrule
 i would be \textcolor{red}{\textbf{eleven}} years old in march , and i had developed strength and skills to rival most \textcolor{red}{boys} my age . & eleven &  age:young (18-44) &  i would be \textcolor{blue}{eighteen} years old in march, and i had developed strength and skills to rival most boys my age. & \textcolor{blue}{I} would be \textcolor{blue}{twenty} years old in \textcolor{blue}{M}arch, and I had developed strength and skills to rival most \textcolor{blue}{men} my age.  \\ % age example (can be noisy)
    \bottomrule
  \end{tabular}
  \end{adjustbox}
  \end{center}
  \caption{Example snippets from \acronym. Annotators selected the `chosen word' as demographic-denoting during the first stage of dataset creation (bolded). Words highlighted in \textcolor{red}{red} in the source appear to be on the coreference chain to the `chosen word' (bold), words highlighted in \textcolor{blue}{blue} were changed by the human or the perturber. }
  \label{tab:datasetexamples}
\end{table*}

In this section, we discuss \dataset, a first-of-its-kind human-annotated dataset of 98,583 text examples we collected for training a controllable generation model to perturb demographic references in text (see examples in \autoref{tab:datasetexamples}). 

\paragraph{Preprocessing:} We sampled the original source data for \acronym~from a range of permissively licensed NLP datasets: BookCorpus~\citep{Zhu_2015_ICCV}, SST~\citep{socher-etal-2013-recursive}, SQuAD~\citep{rajpurkar-etal-2016-squad},  MNLI~\citep{williams-etal-2018-broad} and ANLI~\citep{nie-etal-2020-adversarial}. We elected to use source data from multiple different datasets---ranging from books to sentiment analysis, question answering, and natural language inference---because we want a \perturber~that can perform well regardless of text domain and task. We also sampled Wikipedia articles to a range of snippet lengths up to 20 sentences. For any multi-segment input (for instance, the premise and hypothesis for NLI tasks), we concatenated each input segment, separating them by a special $\texttt{<SEP>}$ token.\footnote{Examples with multiple segments are concatenated with a $\texttt{<SEP>}$ token and fed as a single sequence into the perturber. Then, we mapped the perturbed segments to the original fields to ensure all references are preserved, i.e., for QA, if a person's name was changed in the question, it is changed to the same name in the answer (see \autoref{sec:preservelabel}).}

We first computed a ``perturbability score'' for each source text snippet to determine whether to present it to annotators. We pre-compiled a list of 785 known demographic terms, including names from \newcite{ma-etal-2021-dynaboard}, across gender, race/ethnicity and age demographic attributes. Since word lists are limited in coverage, we also use the Stanza Named Entity Recognition (NER) module \citep{qi-etal-2020-stanza} to identify named entities in text. For each text snippet $s$, we compute

\begin{equation}
%\scriptsize
\fontsize{9.5pt}{9.5pt}
    \texttt{perturbability}(s) = \frac{m_0 \cdot \texttt{NER}(s) + m_1 \cdot |s \cap \mathcal{D}_\text{d}|}{|s|}
    \label{eq:perturbability-score}
\end{equation}

where $m_0$ and $m_1$ are adjustable weights for the Named Entity Recognition (NER) system and word list, and $\mathcal{D}_\text{d}$ denotes the dictionary of terms for demographic axis $d$. Ranking text samples with the perturbability score allows us to filter for examples likely to contain demographic information. This process valued precision over recall: we accepted the fairly high false positive rate, since we employed human annotators to inspect the preprocessed sentences later in data creation, and we excluded snippets that were not perturbable. 

\paragraph{Data Creation:} 524 English-speaking crowdworkers generated \acronym~from preprocessed snippets through a three-stage annotation process (see \autoref{sec:annotationdetails}) executed on Amazon Mechanical Turk (AMT) over the span of 5 months, excluding U.S. public holidays. We paid a competitive hourly wage in keeping with local labor laws. The demographics of our workers roughly matches a recent survey of AMT workers ~\citep{moss-etal-2020-demographic}, which found that workers skew white, and are more likely to be women than men. For a more detailed demographic breakdown, see \autoref{app_sec:annotators}. 

We created task-specific onboarding qualifications for each stage of data collection. % (see \autoref{sec:annotationdetails}). %,  \autoref{fig:datacollectionphase1}, \autoref{fig:datacollectionphase2} and \autoref{fig:datacollectionphase3}). 
In addition to onboarding requirements, we monitored annotators' performance in-flight and assembled an allow-list of 163 high performing annotators to collect more challenging annotations, such as longer Wikipedia passages.

% We utilized the allow-listed annotator pool to collect more challenging annotations, such as long Wikipedia passages, and set up ad hoc communication with them over email. %We thank several of our star annotators in the acknowledgments. 

\if0
\begin{figure*} %placeholder fig. sketch
      \begin{center}
    \begin{adjustbox}{width=150}
    \includegraphics[width=150]{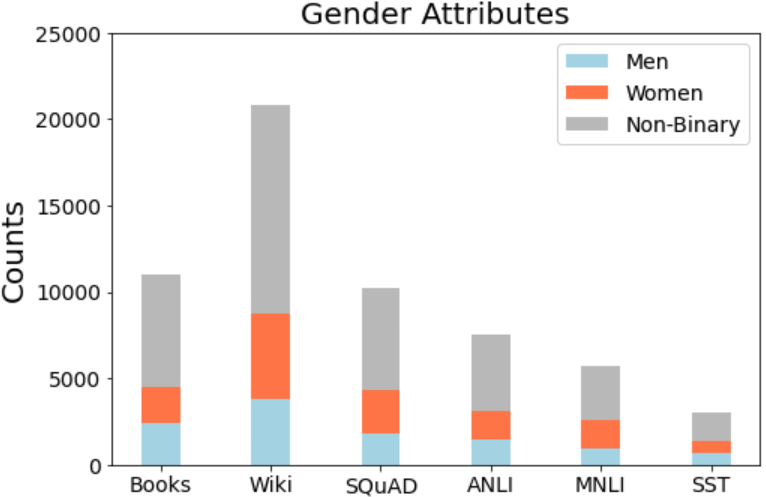}
    \end{adjustbox}
    \begin{adjustbox}{width=150}
    \includegraphics[width=150]{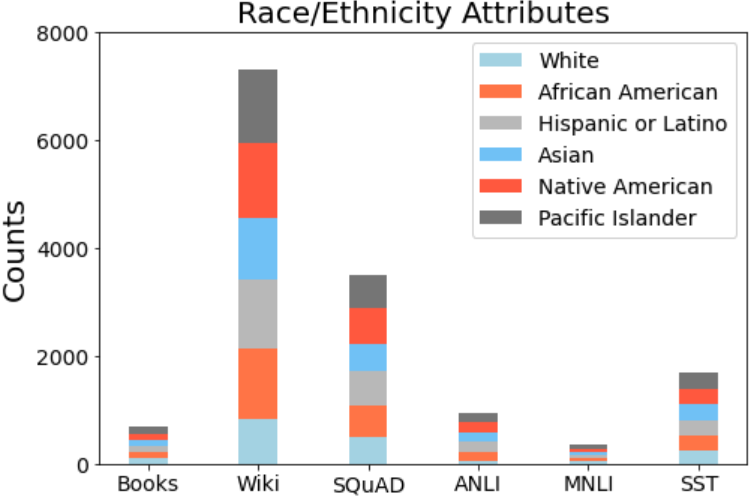}
    \end{adjustbox}
    \begin{adjustbox}{width=150}
    \includegraphics[width=150]{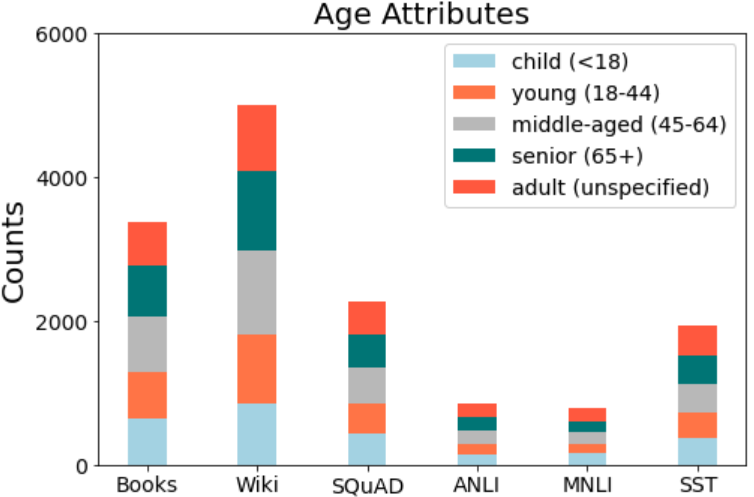}
    \end{adjustbox}
    \end{center}
    \caption{Breakdown of demographic axes and source data types in \acronym. The \textit{x}-axis shows number of examples. Analysis is shown on the rewritten examples.}
    \label{fig:demographic-breakdown}
\end{figure*}
\fi

\pgfplotstableread{
Label           man         woman       non-binary
BC      2402        4468        6505
Wiki       3767        8725        12134
SQuAD           1791        4314        5920
ANLI            1478        3128        4433
MNLI            963         2559        3165
SST-2           699         1340        1646
}\genderdata

\pgfplotstableread{
Label           white       AA          HS      AS      NA      PI
BC      94          122         113     116     117     127
Wiki       822         1300        1276    1147    1399    1360
SQuAD           483         595         628     505     672     604
ANLI            49          179         181     166     184     187
MNLI            38          63          67      50      67      69
SST-2           248         279         284     281     293     293
}\racedata

\pgfplotstableread{
Label           child       young       middle      senior      adult
BC      650         641         768         702         607
Wiki       848         966         1151        1110        922
SQuAD           442         414         484         475         459
ANLI            142         143         192         191         183
MNLI            156         123         183         146         185
SST-2           377         340         412         397         399
}\agedata
\pgfplotsset{scaled x ticks=false}

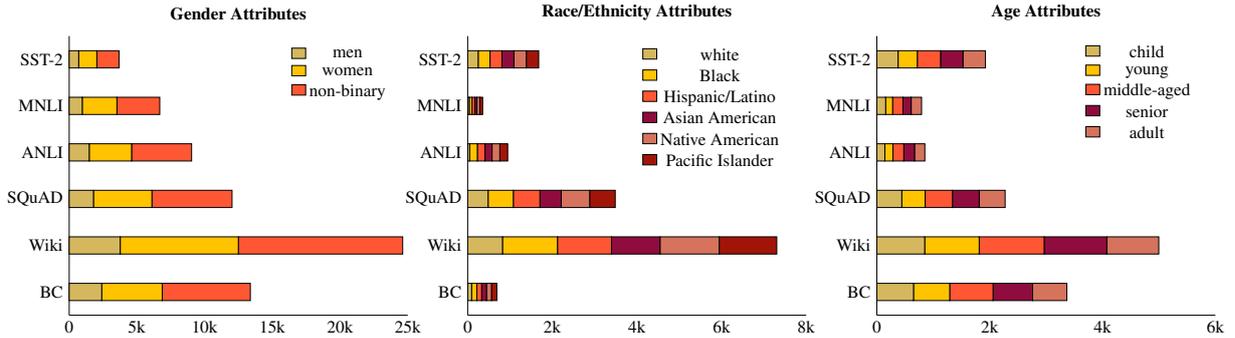
\begin{figure*}
\begin{minipage}[t]{.32\textwidth}
    \scalebox{0.65}{
    \begin{tikzpicture}
        \begin{axis}[
        xbar stacked,
        axis x line*=bottom,
        axis y line*=left,
        xmin=0,xmax=25000,
        xtick={0,5000,10000,15000,20000,25000},
        xticklabels={0,5k,10k,15k,20k,25k},
        ytick=data,     % Use as many tick labels as y coordinates
        extra x tick style = {
            log identify minor tick positions=false,
        }, 
        legend style={at={(axis cs:16000,4)},anchor=south west,draw={none}},
        yticklabels from table={\genderdata}{Label},
        x tick label style={
            /pgf/number format/fixed,/pgf/number format/precision = 0,
            },
        xtick style={draw=none},
        ytick style={draw=none},
        title={\textbf{Gender Attributes}}
        ]
        \addplot [fill=color1] table [x=man, meta=Label,y expr=\coordindex] {\genderdata};
        \addplot [fill=color2] table [x=woman, meta=Label,y expr=\coordindex] {\genderdata};
        \addplot [fill=color3] table [x=non-binary, meta=Label,y expr=\coordindex] {\genderdata};
        \legend{men,women,non-binary}
    
        \end{axis}
    \end{tikzpicture}
    }
 \end{minipage}\hspace*{0.1em}
 \begin{minipage}[t]{.32\textwidth}
    \scalebox{0.65}{
        \begin{tikzpicture}
        \begin{axis}[
        xbar stacked,
        axis x line*=bottom,
        axis y line*=left,
        xmin=0,xmax=8000,
        xtick={0,2000,4000,6000,8000},
        xticklabels={0,2k,4k,6k,8k},
        ytick=data,     % Use as many tick labels as y coordinates
        legend style={at={(axis cs:4000,2.5)},anchor=south west,draw={none}},
        yticklabels from table={\racedata}{Label},
        x tick label style={
            /pgf/number format/fixed,
            },
        xtick style={draw=none},
        ytick style={draw=none},
        title={\textbf{Race/Ethnicity Attributes}}
        ]
        \addplot [fill=color1] table [x=white, meta=Label,y expr=\coordindex] {\racedata};
        \addplot [fill=color2] table [x=AA, meta=Label,y expr=\coordindex] {\racedata};
        \addplot [fill=color3] table [x=HS, meta=Label,y expr=\coordindex] {\racedata};
        \addplot [fill=color4] table [x=AS, meta=Label,y expr=\coordindex] {\racedata};
        \addplot [fill=color5] table [x=NA, meta=Label,y expr=\coordindex] {\racedata};
        \addplot [fill=color6] table [x=PI, meta=Label,y expr=\coordindex] {\racedata};
        \legend{white,Black,Hispanic/Latino,Asian American,Native American,Pacific Islander}
    
        \end{axis}
    \end{tikzpicture}
    }
\end{minipage}\hspace*{0.5em}
\begin{minipage}[t]{.32\textwidth}
\scalebox{0.65}{
    \begin{tikzpicture}
        \begin{axis}[
        xbar stacked,
        axis x line*=bottom,
        axis y line*=left,
        xmin=0,xmax=6000,
        xtick={0,2000,4000,6000},
        xticklabels={0,2k,4k,6k},
        ytick=data,
        legend style={at={(axis cs:3600,3.1)},anchor=south west,draw={none}},
        yticklabels from table={\agedata}{Label},
        x tick label style={
            /pgf/number format/fixed,
            },
        xtick style={draw=none},
        ytick style={draw=none},
        title={\textbf{Age Attributes}}
        ]
        \addplot [fill=color1] table [x=child, meta=Label,y expr=\coordindex] {\agedata};
        \addplot [fill=color2] table [x=young, meta=Label,y expr=\coordindex] {\agedata};
        \addplot [fill=color3] table [x=middle, meta=Label,y expr=\coordindex] {\agedata};
        \addplot [fill=color4] table [x=senior, meta=Label,y expr=\coordindex] {\agedata};
        \addplot [fill=color5] table [x=adult, meta=Label,y expr=\coordindex] {\agedata};
        \legend{child,young,middle-aged,senior,adult}
    
        \end{axis}
    \end{tikzpicture}
    }
\end{minipage}
\caption{Breakdown of demographic axes and source data types in \acronym. `Wiki' refers to Wikipedia and `BC' refers to BookCorpus. The \textit{x}-axis shows number of examples for each attribute. Analysis is shown for the rewritten examples.}
\label{fig:demographic-breakdown}
\end{figure*}

\paragraph{The Dataset:} \acronym~contains 98,583 pairs of original and demographically perturbed text snippets, along with perturbed demographic words and attributes. % (see examples in \autoref{tab:datasetexamples}).
The prevalence of demographic terms from different axes differs by dataset, and the overall percentage of rewrites for each demographic axis are 70.0\% for gender, 14.7\% for race/ethnicity, and 14.6\% for age. The higher prevalence of gender overall is related to the fact that gender is morphologically marked on pronouns in English---which are highly frequent---while age and race descriptors are not. We report the distribution of examples in \acronym~that contain words from a particular demographic axis and attribute in \autoref{fig:demographic-breakdown}.

By design, demographic attributes are roughly balanced in \acronym~within each axis. This is in contrast to other commonly used source datasets which have imbalances across demographic attributes: for example, one estimate suggests that 70\% of gendered sentences in Wikipedia
referred to men \citep{sun-etal-2021-they}, and the training dataset for PaLM, another recently released large language model, had five and a half times as many \textit{he/him} references as  \textit{she/her} ones ~\citep[p.67]{chowdhery-etal-2022-palm}. In short, attributes like `men' and `white' appear to have been more present in the source data and our data collection process perturbs them to other attributes, thereby upsampling rarer demographic attributes. This results in rough attribute parity in \acronym.

To verify that \acronym~is of high quality and that crowdworkers did actually target the correct axes and attributes, four experts performed a preliminary dataset audit inspired by \newcite{blodgett-etal-2021-stereotyping}: from a representative sample of 300 snippets from \acronym, they found the data to be of relatively high quality (see \autoref{tab:data-quality-summary}). We also estimated naive token-wise interannotator agreement to be 95\% (see \autoref{sec:iaa} for more metrics and futher details), suggesting little variation in how crowdworkers perturb.\footnote{Anecdotally, variation arises when multiple rewrites are acceptable. \textit{Queen Victoria} can be perturbed to \textit{King Victor} or to \textit{King Jacob}. Neopronouns \textit{xe, ze, ey}, instead of \textit{they} for non-binary rewrites are also a valid source of variation.} 

\section{Training the Demographic Perturber}

We frame training a demographic perturber as a conditional sequence-to-sequence task. Given input snippet $s$, perturbable word $w$ and target attribute $a_t$, we seek to learn $P(\widetilde{s} | s, w, a_t)$, where $w$ and $a_t$ are discrete control variables that we prepend to perturber inputs. The perturber inputs take the form \texttt{[perturbable word] [target attribute] <PERT\_SEP> [input]}. The \perturber~is a finetuned BART model \citep{lewis-etal-2020-bart} with 24 layers, 1024 hidden size, 406M parameters, and 16 attention heads. To train the \perturber, we finetune BART on \acronym~using the ParlAI library\footnote{\href{https://github.com/facebookresearch/ParlAI}{\texttt{github.com/facebookresearch/ParlAI}}} \citep{miller-etal-2017-parlai}, with training parameters provided in \autoref{tab:perturber-training-params}. We achieve a BLEU score of 88.0 (measured against the source) on the validation set, and perplexity of 1.06, which is likely low because perturbation preserves the majority of tokens.

Perturbing large ML training datasets is an important application of perturbation augmentation. Therefore, it is crucial that generation is fast and scalable to large text corpora. We experimented with different architectures and generation techniques to optimize for both quality and efficiency. Notably, T5 \cite{raffel-etal-2020-exploring} performed slightly better on certain NLP metrics (such as BLEU-4), but used much more memory during training and inference, resulting in \textit{16x} slower generations in a distributed setting. We also explored different ways of decoding, and surprisingly, found that greedy decoding performs as well as beam search in our setting. We therefore use greedy decoding in our perturbation augmentation applications, which is also memory efficient.

\begin{table}[t]
  \begin{center}
  %\begin{adjustbox}{width=\linewidth}
  \begin{tabular}{p{1.3cm}wr{1.2cm}wr{2.1cm}wr{1.4cm}}
    \toprule
    %\multicolumn{1}{c}{}                   \\
     & \bf BLEU & \bf Lev. Distance & \bf ROUGE-2 \\
    \midrule
    \perturber & \textbf{86.7}  & \textbf{5.20} & \textbf{90.9}     \\
    AugLy     & 80.6 & 7.88   & 87.2   \\
    TextFlint     & 72.3       & 9.25  & 78.3\\
    \bottomrule
  \end{tabular}
  %\end{adjustbox}
  \end{center}
  \caption{Our learned \perturber\ matches human-written perturbations better than heuristic perturbations do.}
  \label{tab:heuristic-perturber}
\end{table}

\begin{figure}
\centering
\begin{tikzpicture}
    \node{
    \parbox{20em}{
        \textbf{[Original]} she bent over to kiss her friends cheek before sliding in next to her .\\
        \textbf{[Perturber]} \yellowemph{He} bent over to kiss \yellowemph{his} friends cheek before sliding in next to her .\\
        \textbf{[AugLy]} \blueemph{he} bent over to kiss \blueemph{him} friends cheek before sliding in next to \blueemph{him} .\\
        \textbf{[TextFlint]} she bent over to kiss her friends cheek before sliding in next to her .
    }
    };
\end{tikzpicture}
\caption{Examples perturbed with heuristic approaches (AugLy and TextFlint), or the \perturber~(changed words highlighted); TextFlint did not perturb any words.}
\label{fig:heuristics-viz}
\end{figure}

\paragraph{Comparison to Heuristics.}\label{sec:heuristics} Is it necessary to train a \perturber, or can we just use heuristics? Previous approaches relied on word lists \citep{zhao-etal-2019-gender} or designing handcrafted grammars to generate perturbations~\citep{zmigrod-etal-2019-counterfactual, ma-etal-2021-dynaboard, renduchintala-williams-2022-investigating, papakipos-etal-2022-augly}. However, word list approaches are necessarily limited~\citep{dinan-etal-2020-queens} and which words are included can really matter \citep{sedoc-ungar-2019-role}. For instance, attributes are often excluded for being hard to automate: e.g., \textit{Black, white} have been excluded because they often denote colors in general \citep{ma-etal-2021-dynaboard}. Grammar-based approaches also require ad hoc solutions for phonological alternations (\textit{\textbf{a} banana} v.~\textit{\textbf{an} apple}), and struggle with one-to-many-mappings for pronouns \citep{sun-etal-2021-they}, often incompletely handling pronoun coreference chains. We find that a neural \perturber~trained on high quality human annotations can correctly identify perturbable words and their coreference chains, and then generate rewritten text that is grammatical, fluent and preserves overall meaning.

We compare the \perturber~to several state-of-the-art heuristic-based systems on a human annotated evaluation set, and find that the \perturber~consistently outperforms heuristic alternatives. The \perturber~generations show higher BLEU~\citep{papineni-etal-2002-bleu} and ROUGE scores than do AugLy~\citep{papakipos-etal-2022-augly} and TextFlint \citep{wang-etal-2021-textflint}, as well as lower Levenshtein distance\footnote{Distance was modified to compute word-level distance.} to the human generated perturbations~(see \autoref{tab:heuristic-perturber}).

Qualitatively, we observe that the \perturber~generally outputs intelligent, human-like text rewrites. \autoref{fig:heuristics-viz} shows an example in which the \perturber~correctly inflects the pronoun ``his'', whereas heuristics failed. We additionally find that the \perturber~is capable of perturbing complex passages, such as the first example in \autoref{tab:datasetexamples}, where the \perturber~changed nouns, pronouns, and names referring to the selected entity, while maintaining fluency and coherence.

\section{Results}\label{sec:results}

We present results showing that using the \perturber~leads to fairer models during pretraining (\autoref{sec:fairberta}) and to fairer models during finetuning without sacrificing accuracy (\autoref{subsec:classification}).

\subsection{\fairberta: Perturbation Augmented Pretraining}\label{sec:fairberta}

\begin{table*}[t!]
  \begin{center}
\normalsize
  \begin{tabular}{llrrp{.1em}rr}
  %\cmidrule{3-6}
  
\toprule
   & & \textbf{RoBERTa}
   & \textbf{\fairberta}
   & &
   \textbf{RoBERTa$^\dagger$}
   & \textbf{\fairberta} \\    
   & &
   \multicolumn{2}{c}{\it 16GB of training data} &
   & \multicolumn{2}{c}{\it 160GB of training data} \\
    %\multirow{1}{*}{\bf Dataset} & \multirow{1}{*}{\bf Axis} & & & & & & & &\\
    \toprule
    \multirow{3}{*}{HolisticBias} 
    & \it gender &  36.1 & \textbf{19.9} &
    & 40.6 & \bf 35.7 \\
    
    & \it race  & 27.3 & \textbf{23.8} &
     & 28.4 & \bf 27.6 \\
    
    & \it age & 42.9 & \textbf{38.9} &
    & \textbf{36.4} & 41.7 \\
    
   %&  \it mean & \textbf{52.9} & \textbf{52.6} & 55.4 & 56.7 & 58.5 & 58.5 & \textbf{57.3} & \textbf{56.7} \\
    
    \midrule
     \multirow{1}{*}{\textsc{Weat}/\textsc{Seat}}
    & \it \% sig. tests & 53.5 & \bf 40.0 &  & 60.0 & \bf 36.7  \\
    
    % & \it effect size & \bf 0.58 & 0.73 & & 0.89 & \bf 0.79\\
     \midrule
    
    \multirow{ 3}{*}{CrowS-Pairs} 
    & \it gender    & 52.3 & \textbf{51.9} &
    & 55.0 & \textbf{51.5} \\
    
    & \it race &   \textbf{55.0} & \textbf{55.0} &
    & \textbf{53.9} & 57.6 \\
    
    & \it age  & \textbf{50.6} & 63.2 &
    & 66.7 & \textbf{63.2} \\
    
    %& \it mean & 0.30 & 0.35 & \textbf{0.27} & \textbf{0.28} & 0.35 & \textbf{0.35} & \textbf{0.21} & 0.38\\
     
   \bottomrule
    
  \end{tabular}
  \end{center}  
  \caption{Results of \fairberta~and RoBERTa on 3 fairness metrics across varying training dataset sizes. Numbers are percentages of metric tests revealing bias. RoBERTa$^\dagger$ refers to the model from \newcite{liu-etal-2019-roberta}; all other models were trained from scratch. For CrowS-Pairs, closer to 50 means a more fair model; for WEAT/SEAT \& HolisticBias, lower means more fair. See \autoref{sec:fairberta} for more details.
  }
  \label{tab:fairnessdatasets}
\end{table*}

% We find that LMs pretrained on data that has been perturbation augmented not only show improvements according to current state-of-the-art fairness metrics, but they also perform as well on downstream tasks.

\paragraph{\textit{Setting:}} We train \fairberta~with the RoBERTa$_{\textrm{\textsc{Base}}}$ architecture~\citep{liu-etal-2019-roberta} using 256 32GB V100 GPUs for 500k steps. To generate training data for \fairberta, we apply the \perturber~to the RoBERTa training corpus \citep{liu-etal-2019-roberta} to help balance the representation of underrepresented groups (see \autoref{fig:demographic-breakdown}) and thereby reduce the prevalence and severity of unwanted demographic associations. During perturbation augmentation, we sample contiguous sequences of 256 tokens and select a demographic word and target attribute with uniform probability, which are provided as inputs to the \perturber. %Passages that do not have perturbable references are left unchanged, preserving the size of the training corpus. 
Although it would be in principle straightforward to upsample the training data size appreciably, keeping data size fixed allows us to make a direct comparison between \fairberta~and RoBERTa on a variety of fairness metrics and downstream tasks. We train \fairberta~and RoBERTa on the full RoBERTa training corpus (160GB) and the BookWiki subset (16GB), and show that our observations on fairness and accuracy are consistent.

\paragraph{\textit{Fairness Evaluations:}} We compare \fairberta~to RoBERTa trained with the same settings according to their performance on three fairness evaluation datasets. For CrowS-Pairs \citep{nangia-etal-2020-crows}, we report the percentage of examples for which a model assigns a higher
(pseudo-)likelihood to the stereotyping sentence over the less stereotyping sentence. %\footnote{Although many of the fairness metrics that are standard in NLP have flaws \citep{blodgett-etal-2021-stereotyping}, we unfortunately have few alternatives.}
For the template-based Word Embedding Association~Test~(\textsc{Weat}, \citealt{caliskan-etal-2017-semantics}) and Sentence Encoder Association Test~(\textsc{Seat}, \citealt{may-etal-2019-measuring}), we report the percentage of statistically significant tests and their average effect size. Lastly, % a new, larger bias measurement dataset,
for HolisticBias (HB, \citealt{smith-etal-2022-im}), we measure the percentage of pairs of descriptors by axis for which the distribution of pseudo-log-likelihoods~\citep{nangia-etal-2020-crows} in templated sentences significantly differs. %, focusing on gender, race/ethnicity, and age.

\begin{table*}
  \begin{center}
\begin{adjustbox}{width=\linewidth}
\addtolength{\tabcolsep}{0pt}
\renewcommand{\arraystretch}{1.0}
\tiny
\begin{tabular}{lcrrrrrrrrr}
    \toprule
    \bf Model & \bf Tuning & \bf Size & \bf CoLA & \bf SST-2 & \bf STS-B & \bf QQP & \bf RTE & \bf QNLI & \bf Avg.\\
     \midrule
    \fairberta & orig. & 16GB & 62.81 & 92.66 & 88.37 & 91.22 & 72.75 & 92.13 & 83.32 \\
    RoBERTa & orig. & 16GB & 59.81 & 93.92 & 89.87 & 91.17 & 72.92 & 91.89 & 83.26 \\ \midrule
    \fairberta & orig. & 160GB & 61.57 & 94.61 & 90.40 & 91.42 & 76.90 & 92.99 & 84.65 \\
    RoBERTa$^{\dagger}$ & orig. & 160GB & 61.36 & 93.50 & 90.90 & 91.77 & 75.50 & 92.70 & 84.29 \\
    \midrule\midrule
    \fairberta & fair & 16GB & 61.37 & 92.20 & 87.64 & 90.93 & 70.03 & 92.13 & 82.38 \\
    RoBERTa & fair & 16GB & 58.09 & 93.58 & 88.66 & 91.04 & 71.12 & 91.73 &  82.37  \\ \midrule
    \fairberta & fair & 160GB & 60.60 & 94.95 & 89.63 & 91.49 & 75.09 & 92.77 & 84.09 \\
    RoBERTa$^{\dagger}$ & fair & 160GB & 59.71 & 93.50 & 90.20 & 91.56 & 75.80 & 92.70 & 83.91 \\ 
    \bottomrule
  \end{tabular}
    \end{adjustbox}
  \end{center}
    \caption{\fairberta~matches RoBERTa in Downstream Task Accuracy (GLUE Benchmark). Tuning refers to whether models are finetuned on original datasets or ``fairtuned'' on perturbed ones (denoted with `fair'). RoBERTa and \fairberta~models report similar accuracy regardless of training size and tuning approach. We report Matthew's correlation for CoLA, Pearson's correlation for STS-B, and accuracy for all other tasks. Results are the median of 5 seeded runs.
    A dagger marks the \citeauthor{liu-etal-2019-roberta} model.}
  \label{tab:finetuneacc}
\end{table*}

\paragraph{\textit{\fairberta~is more fair:}} Overall, \fairberta~shows improvements in fairness scores over training-size-matched RoBERTa models across our evaluations, and across two training dataset sizes (see \autoref{tab:fairnessdatasets}). %FairBERTa models outperform RoBERTa models according to HolisticBias and WEAT/SEAT metrics, although the CrowS results are more equivocal. 
\fairberta~models show reduced demographic associations overall across HB templates, and have notably fewer statistically significant associations on \textsc{Weat}/\textsc{Seat}. CrowS-Pairs is more equivocal: e.g., \fairberta~(16GB) is closer than RoBERTA (16GB) to the desired score of 50\% (demographic parity) for gender, but not for age. Worse performance on the age category is possibly due to the varied ways in which age is conveyed in language, e.g., \textit{I was born 25 years ago} vs. \textit{I am a child}. While the \perturber~is capable of perturbing phrases with numbers such as \textit{eleven years old}, general issues with numerical reasoning \citep{dua-etal-2019-drop, geva-etal-2020-injecting, lin-etal-2020-birds} may still be present. %We leave the construction of a more sophisticated system for demographic identification of age to future work.

We find that fairness metrics sometimes report conflicting results, corroborating other recent findings \citep{delobelle-etal-2021-measuring, goldfarb-tarrant-etal-2021-intrinsic}. While \textsc{Weat}/\textsc{Seat} tests and HB evaluation find \fairberta~(160GB) to be more fair along the race axis, CrowS-Pairs reported a better score for RoBERTa (160GB). Inconsistencies may be partly explained by data noise in CrowS-Pairs \cite{blodgett-etal-2021-stereotyping}, but we believe that the agreement (or lack thereof) of different NLP bias measurements warrants further exploration, and closer examinations of fairness evaluation datasets.

\paragraph{\textit{\fairberta~has no Fairness-Accuracy Tradeoff:}} Previously, a fairer model often meant accepting lower task performance \citep{zliobaite-2015-relation,pmlr-etal-menon-2018} or seeking a Pareto optimal solution \citep{berk-etal-2017-convex, zhao-gordon-2019-inherent}. %Past work in algorithmic fairness has found evidence that for certain settings, such as binary classification 
%\citep{zliobaite-2015-relation,pmlr-etal-menon-2018}, to get a fairer model you may have to accept a less accurate one %or you can seek a pareto optimal solution \citep{berk-etal-2017-convex, zhao-gordon-2019-inherent}.
To determine whether there is a tradeoff between downstream task accuracy and fairness in our setting, we evaluate on 6 GLUE benchmark tasks~\cite{wang2018glue}: sentence acceptability~\citep[CoLA]{warstadt-etal-2019-neural}, sentiment analysis~\citep[SST-2]{socher-etal-2013-recursive}, text similarity~\citep[STS-B]{cer-etal-2017-semeval}, textual entailment~\citep[RTE]{dagan-etal-2005-pascal, haim-etal-2006-second, giampiccolo-etal-2007-third, bentivogli-etal-2009-fifth}, and question answering~\citep{rajpurkar-etal-2016-squad} recast to textual entailment (QNLI)%semantic equivalence over questions\footnote{\href{data.quora.com/First-Quora-Dataset-Release-Question-Pairs}{data.quora.com/First-Quora-Dataset-Release-Question-Pairs}}
.\footnote{We exclude several GLUE tasks for which the number of demographically perturbable examples was too low to draw meaningful conclusions. We follow \newcite{liu-etal-2019-roberta}'s training procedure, conducting a limited hyperparameter sweep for each task varying only learning rate and batch size. For each task, we finetune for 10 epochs and report the median development set results from five random initializations.}

\fairberta~models match the performance of RoBERTa models trained under the same setting to within 0.40\% accuracy on average (see top half of \autoref{tab:finetuneacc}). For some tasks (CoLA, SST-2, RTE and QNLI), \fairberta~(160GB) also slightly outperforms RoBERTa (160GB) and averages 0.75\% higher overall accuracy on these tasks. 

% The largest drop in task accuracy occurs for RTE, where RoBERTa (160GB) had \% higher accuracy than \fairberta~(160GB). 

% These early findings are promising and support the claim from \citep{wick-etal-2019-unlocking} that ``fairness and accuracy are sometimes in accord''. We believe our work has exciting implications for the practice of pretraining large language models, and that it will inspire more exploration of demographic diversity in model training corpora.

\begin{table*}[ht]
  \begin{center}
  \normalsize
  \begin{tabular}{lcrrrrrrc}
    \toprule
   \bf Model & \bf Tuning & \bf Size &\bf CoLA &  \bf SST2 &  \bf QQP &  \bf RTE & \bf QNLI &  \bf Avg.\\
   \midrule
    \fairberta & orig. & 16GB & 5.46 & 2.04 & 5.61 & \bf 6.45 & \bf 1.70 & 4.25  \\
    \fairberta & fair & 16GB & \bf 4.20 & \bf 1.02 & \bf 3.34 & \bf 6.45 & 1.94 & \bf 3.39 \\\midrule
    \fairberta & orig. & 160GB & 5.88 & 1.02  & 5.56 & \bf 3.23 & 2.17 & 3.57  \\
    \fairberta & fair & 160GB & \bf 4.41 & \bf 0.51 & \bf 2.86 & 6.45 & \bf 1.70 & \textbf{3.19} \\
    
    \midrule \midrule
    RoBERTa & orig. & 16GB & 6.51 & \bf 1.02 & 6.89 & \bf 6.45 & 2.88 & 4.75 \\
    RoBERTa & fair & 16GB & \bf 5.46 & 3.06 & \bf 3.43 & 6.86 & \bf 1.58 & \textbf{4.08}  \\ \midrule
    RoBERTa$^{\dagger}$ & orig. & 160GB & 6.93 & 2.55 & 7.60 & \bf 4.03 & 2.17 & 4.66 \\
    RoBERTa$^{\dagger}$ & fair & 160GB & \bf 3.78 & \bf 1.02 & \bf 3.22 & 6.45 & \bf 1.67 & \textbf{3.23} \\
    \bottomrule
  \end{tabular}
  \end{center}
    \caption{The \fairscore~for fairtuned models is lower in general. A lower \fairscore{}, i.e., the percentage of classifier predictions that change during inference for a single model between the original evaluation set and the same evaluation set after perturbation augmentation, corresponds to a fairer model. %We finetune and fairtune RoBERTa$_{\textrm{BASE}}$ and fairBERTa$_{\textrm{BASE}}$;
    The lowest \fairscore{} for each task and setting is bolded.  RoBERTa$^{\dagger}$ is the model from \newcite{liu-etal-2019-roberta}.
    }
  \label{tab:fairscoreresults}
\end{table*}

\subsection{Fairtuning: Finetuning on Perturbed Data}\label{subsec:classification}

%\paragraph{Setting:} One of the prevailing methods for training NLP classifiers relies on finetuning language models that were pretrained on large scale datasets. Given the cost of pretraining in terms of both time and resources, our approach in Section~\ref{sec:fairberta} may not be practical in all scenarios for reducing bias. For this reason, we also introduce \textbf{fairtuning}, an approach for automatically debiasing models during finetuning which is related to counterfactual generation \citep{wu-etal-2021-polyjuice}.
\paragraph{\textit{Setting:}} In addition to comparing downstream performance in a traditional finetuning setting, we also compare performance and fairness during \textbf{fairtuning}, where models are finetuned on demographically perturbed downstream datasets (see \autoref{sec:approach}). The number of perturbable examples and the proportions of demographic axes varies across fairtuning data by task (see statistics in \autoref{tab:glue-stats}, and examples in \autoref{tab:glue-examples}).

\paragraph{\textit{Fairtuning does not degrade downstream task accuracy:}} Fairtuned models match their finetuned counterparts in accuracy on the original (unperturbed) GLUE validation sets (compare the top half of \autoref{tab:finetuneacc} to the bottom). Surprisingly, for some tasks (SST-2, QQP and RTE), fairtuning resulted in slightly higher original validation set performance than finetuning does for some model configurations. The largest drop in performance from fairtuning occurs for RTE, where \fairberta~trained on BookWiki (16GB) shows a decrease of 2.72\% in accuracy. Swings on RTE may be due to its smaller size (see \autoref{tab:glue-stats}), as we observe more variance across finetuning runs as well. Finetuning or fairtuning from an existing NLI checkpoint, as in \citealt{liu-etal-2019-roberta}, might result in more stability.

% or high percentage of perturbed examples (see \autoref{tab:glue-stats}), as we observe smaller swings for larger, less perturbable datasets (QQP, QNLI, SST-2). 

%Averaged across all tasks, models fairtuned on perturbed data differ by less than 1 percent from models finetuned on original data, suggesting that fairtuning does not significantly degrade downstream task performance; in some cases, it may even improve task accuracy. 
\subsection{Measuring Fairness with the \Fairscore} 

\paragraph{\textit{Setting:}} Finally, we compute the \textbf{\fairscore} as an extrinsic fairness evaluation metric. Recall that, given a classifier and evaluation set, the \fairscore~of the classifier is the percentage of predictions that change when the input is demographically altered with the perturber. 

\paragraph{\textit{\Fairscore~is best for Fairtuned Models:}} Fairtuned models have lower (i.e., better) fairscores on average\footnote{We report on all tasks except STS-B, a regression task, because the fairscore is defined for classification tasks.}, meaning that their predictions change the least from perturbation (see \autoref{tab:fairscoreresults}). On average, fairtuned models saw a 0.84 point reduction in the \fairscore~as compared to models finetuned on unperturbed data; this is true for both RoBERTa and \fairberta~and across training data sizes. We also find that \fairberta~models are more robust to demographic perturbation on downstream tasks, even when finetuned on the original datasets (\autoref{tab:fairscoreresults}). \fairberta~models have lower \fairscore s than RoBERTa models pretrained on similar sized datasets.

%In some cases, we observe an additive effect where models that are both pretrained \emph{and} finetuned on demographically perturbed data show more robustness to demographic perturbation on downstream tasks. Notably, for 2 out of 5 tasks (QQP, SST-2), the lowest \fairscore~across all models is reported by \fairberta~(160GB) fairtuned on demographically altered data.
We also observe an additive effect where models that are both pretrained \emph{and} finetuned on demographically perturbed data show more robustness to demographic perturbation on downstream tasks. Notably, the fairtuned versions of \fairberta~(16BG) and \fairberta~(160GB) have better average \fairscore s in general.
The fairtuned \fairberta~(160GB) model reports the lowest average \fairscore~across all tasks (3.19). In our setting, we do not observe any relationship between demographic bias and data size in downstream tasks, suggesting that models of any size can learn demographic biases.

Overall, we find that perturbation augmentation can mitigate demographic bias during classification without any serious degradation to task performance for most tasks on the GLUE benchmark (see \autoref{tab:finetuneacc}). While we do observe an interesting additive effect where LMs are more robust to demographic differences when they are pretrained on demographically altered datasets then fairtuned, we believe that further work is needed to better understand exactly how bias is learned and propagated during different stages of language model training.

\section{Conclusion}
As language models become more powerful and more popular, more attention should be paid to the demographic biases that they can exhibit. Models trained on datasets with imbalanced demographic representation can learn stereotypes such as \textit{women like shopping}. While recent works have exposed the biases of LMs using a variety of techniques, the path to mitigating bias in large scale training datasets is not always clear. Many approaches to correct imbalances in the dataset have used heuristic rules to identify and swap demographic terms. We propose a novel method that perturbs text by changing the demographic identity of a highlighted word, while keeping the rest of the text the same. We find that our \perturber~model creates more fluent and humanlike rewrites than heuristics-based alternatives. We also show that training on demographically perturbed data results in more fair language models, in both pretrained language models and in downstream measurements, without affecting accuracy on NLP benchmarks. We hope our contributions will help drive exciting future research directions in fairer NLP.

\section{Broader Impact}

\paragraph{Fairwashing:} One of the primary worries when releasing a new method for measuring and improving fairness is the possibility that others will use your methods to make blanket proclamations about the fairness of models without acknowledging the limitations and blindspots of the method. In particular, it is possible that users of our models might infer from our naming conventions (i.e., \fairberta, \fairscore) that our models ought to be deemed ``fair'' or that models performing well according to our metrics ought to be deemed ``fair''. We would like to caution against such an interpretation. Our models appear to be more fair than previous models, but that by no means guarantees they are completely ``fair.'' Researching fairness and bias in NLP data and models is a process of continual learning and improvement and we hope our contributions will help open new avenues that may support the training of even better models in the future.

\paragraph{Factuality:} We have shown above that our augmentation process can sometimes create nonexistent versions of real people, such as discussing an English King Victor (not a historical figure), as opposed to a Queen Victoria (a historical figure). We embrace the counterfactuality of many of our perturbations\footnote{Several of our annotators and some early readers asked about body part terms, as our data collection procedure would have annotators perturb gender references while leaving the body references unchanged, as would our learned perturber model. We have left these examples in the dataset to be inclusive of transgender individuals, and we note that, based on anecdotal samples, these examples are rare.}, but the lack of guaranteed factuality means that our approach may not be well-suited to all NLP tasks. For example, it might not be suitable for augmenting misinformation detection datasets, because peoples' names, genders, and other demographic information should not be changed. For tasks that rely on preserving real world factuality, it would be interesting to explore ways to teach models not to perturb demographic references to known entities, perhaps by relying on a pipeline that includes entity linking systems. That being said, the perturber is fairly general purpose and can perturb text from a wide range of domains. Approaching the problem from a counterfactual angle also means we can imagine follow-up experiments that vary the mix of different demographic characteristics (see \autoref{fig:demographic-breakdown}). One could train a model where all the human references are to a historically underrepresented group (e.g., women) and explore what changes take place in the model's internal representations.

% \paragraph{General Issues with fairness measurement:} We have shown in \autoref{sec:fairnessmetric} that our according to several fairness datasetsCROW-S \cite{nangia-etal-2020-crows}, %StereoSet \cite{nadeem-etal-2021-stereoset}, \cite{parrish-etal-2021-bbq}, Dellobelle survey biased rulers \cite{delobelle-etal-2021-measuring})
    
% There are issues with crowdsourcing stereotypes: Norwegian Salmon \cite{blodgett-etal-2021-stereotyping}

\paragraph{Breaking Statistical Associations:} Our approach weakens statistical associations with demographic axes and attributes regardless of the identity of a particular axis or attribute or the content of the particular association. Which associations are present depends on the source data: if the source data contains more gendered references to men (see \autoref{tab:glue-stats}), this will be balanced in our approach by upsampling references to women and non-binary people (see \autoref{fig:demographic-breakdown}). However, each attribute will get perturbed in the same way, and no associations will be ``spared''. Stereotypical associations will be weakened, but so will non-stereotypical ones. Occasionally, there are associations that some may argue that we don't want to weaken. Because deciding which associations are harmful is often subjective, requiring training and/or lived experience, we have approached this process from a somewhat basic starting point (i.e., weakening all associations), but it would be interesting (and important) to explore more targeted methods for only weakening associations that are known to be harmful. 

\paragraph{Perturbation Augmentation for Hate Speech Detection:} We have motivated the \fairscore~as a relatively task-neutral and scalable way to measure fairness across different types of classification tasks. However, this approach is not a good fit for every possible classification task: for example, certain definitions of hate used for hate speech detection define it as being targeted at particular groups that are minoritized \citep{waseem-etal-2017-understanding}, whereas others define it as against demographic categories that are, often legally, protected \citep{rottger-etal-2021-hatecheck}. The \fairscore~metric, as a simple difference, doesn't distinguish between a difference that harms a majority group from one that harms a minority---in this way, the metric is based on equality, not equity. In short, if we take a hate speech detection example which is labelled as ``hateful'' and pertains to women, if we perturb the example to pertain to men, it may no longer count as ``hateful'' (under definitions that rely on minoritized group status). %as a prerequisite for a negative statement to count as hateful. 
We caution researchers working on tasks like hate speech detection to be careful in considering whether a fairness metric like \fairscore~is appropriate for their use case before they proceed. Also see \autoref{sec:preservelabel} for other task-related complications that one should consider when applying the \fairscore~metric to new tasks. 

Moreover, \newcite{sen-etal-2022-counterfactually} showed that BERT models trained on counterfactually augmented training data to detect hate speech show higher false positive rates on non-hateful usage of identity terms, despite higher out-of-domain generalization capabilities. Despite the fact that they focus on BERT in a slightly different setting, their results still suggest that counterfactually altering the training data might have unforeseen consequences for hate speech, toxic or abusive language detection.

\paragraph{Pronouns can have many-to-many mappings between form and function, and other Linguistic Complications for the Label ``Non-binary'':} Another feature of this work pertains to our label ``non-binary or underspecified''. We are well aware of the fact that gender neutrality and non-binarity are not synonymous. We grouped the two together because many non-binary examples are ambiguous in referring to either (i) an individual who is known to be non-binary, or (ii) an individual whose gender is not specified, or (iii) a plurality of entities \citep{ackerman-2019-syntactic}. This is due in part to the grammatical property of English that the most commonly used non-binary pronoun---singular \textit{they}---is syncretic with the plural pronoun \textit{they}.\footnote{Syncretism refers to the linguistic phenomenon when functionally distinct occurrences of a single word (lexeme, or morpheme) have the same form.} For example, in \textit{every teacher loves their students}, it could be that the speaker knows that the relevant teachers (given the context) are all non-binary, or it could be that the speaker is choosing not to reference the teachers by gender for some reason (the speaker may not know the teachers' genders, they may be a mixed group, or the speaker may just not find gender to be relevant). 

Relatedly, some examples in our dataset maintain the fact that the gender of the perturbed entity is known as a result of quirks of morphological marking in English\footnote{For the perturbation pair from \dataset~\textit{he spun to his feet once more, only to find the girls second dagger pressed against his throat.}$\rightarrow$\textit{they spun to their feet once more, only to find the girls second dagger pressed against their throat.}, we know the output sentence specifies gender only because \textit{throat} is morphologically marked for singular, and generally humans only have one throat. If it were \textit{throats}, we might conclude that \textit{they} is morphologically plural and refers to multiple people of mixed,  unknown, or known non-binary gender.}, but many other examples become ambiguous in this way when we perturb to the `non-binary' attribute. %: \textit{The Prophet took this rock with her to paradise.}$\rightarrow$\textit{The Prophet took this rock with \textbf{them} to paradise.} 
Future work will include a more specific analysis of the examples which were perturbed to non-binary or underspecified to quantify the extent of this ambiguity. Such an analysis project should also explore the use of neopronouns in the dataset.

\section{Limitations}\label{sec:limitations}

\paragraph{Selecting Demographic Categories:} One clear limitation of this work is its reliance on selected categories to perturb. Whenever one categorizes, particularly in the context of social categories, one is excluding some groups, reifying others, and/or making social distinctions that don't fit everyone \citep{keyes-2019-counting}. For example, we rely on US Census categories to delimit possible race/ethnicity attributes, but the Census has numerous shortcomings, including the contentiousness of the Census classification for people of Arab descent as ``white'' \citep{kayyali-etal-2013-us, beydoun-etal-2015-boxed}. 

Another related limitation is the fact that intersectional identities are not a primary focus of this work, which is a clear limitation \cite{buolamwini-gebru-2018-gender}. We observe some coverage of intersectional identities in \acronym, for example names that connote both ethnic and gender identities, and words such as ``grandmother'' that identify gender as well as age. The reason we have left this important topic to future work is that the source data commonly used to train LMs is sorely lacking in references to entities that make explicit all of their identity characteristics. This means that trying to upsample the representation of intersectional identities in text would require injecting attributes, which comes with its own complications; see \citealt{blodgett-etal-2021-stereotyping} for a discussion of the complexity of relevant pragmatic factors, and \citealt{bailey-etal-2022-based} for a gender-related example. Therefore, we feel that entities with multiple identity references need more attention than we could give here. Once we determine how best to handle injecting references with multiple identity attributes, we can also focus on perturbing multiple demographic attributes at once, or perturbing the demographic attributes of multiple entities at once. 

\paragraph{Biases from Data Sourcing:} For this work, we sourced our annotated data from a range of sources to ensure: (i) permissive data licensing, (ii) that our perturber works well on NLU classification tasks for the \fairscore~application, and (iii) that our perturber can handle data from multiple domains to be maximally useful. However, we acknowledge that there may be other existing biases in \acronym~as a result of our data sourcing choices. %\footnote{We thank Yacine Jernite (personal communication) in particular for helping to clarify our thinking around this issue.} 
For example, it is possible that data sources like BookWiki primarily contain topics of interest to people with a certain amount of influence and educational access, people from the so-called ``Western world'', etc. Other topics that might be interesting and relevant to others may be missing or only present in limited quantities. The present approach can only weaken associations inherited from the data sources we use, but in future work, we would love to explore the efficacy of our approach on text from other sources that contain a wider range of topics and text domain differences. 

\paragraph{Crowdsourcing Tradeoffs:} In this work, we relied on crowdworkers to generate perturbations. While human-written perturbations are generally of high quality with respect to grammar, they include data issues such as typos, and can reflect individual preconceptions about what is appropriate or acceptable. Moreover, crowdworker preconceptions may conflict or not be compatible with each other \citep{talat-etal-2021-word}. Take for example the crowdworkers' notion of what counts as appropriate demographic terms. For example, we have observed the use of \textit{Blacks} as a term to refer to ``Black people'' in the final example in \autoref{tab:datasetexamples}. This manner of reference is contested by some, including, for example, the style guidelines from the 501c3 non-profit the National Association of Black Journalists (NABJ)\footnote{\href{https://www.nabj.org/page/styleguideA}{\texttt{www.nabj.org/page/styleguideA}}}, which suggests that journalists should ``aim to use Black as an adjective, not a noun. Also, when describing a group, use Black people instead of just `Blacks.''' We encouraged annotators to use these conventions, but they are unlikely to be uniformly applied, as human group references are prone to change over time~\citep{smith-etal-1992-changing, galinsky-etal-2003-reappropriation, haller-etal-2006-media, zimman-etal-2020-we}. 

Another limitation of our work that relates to crowdsourcing pertains to the perturbation of names. Annotators in the first stage of annotation used their judgment to identify names they believed contain information related to a particular demographic attribute. However, most names can be and are held by people from various genders or racial/ethnic backgrounds, and these proportions are a field of study in themselves \citep{tzioumis2018demographic}. Take an example instance of the open ended rewrite portion of our annotation pipeline that asked the annotator to change \textit{Malcolm} to \texttt{race/eth:african-american}. Some annotators may interpret this to mean that  \textit{Malcolm} doesn't already refer to an Black person, although it might. We accepted annotations in these cases where annotators kept the name unchanged (i.e., the annotator assumed \textit{Malcolm} to refer to an Black person, so the snippet needed no perturbation) and annotations that changed the name (i.e., the annotator assumed \textit{Malcolm} referred to someone from a different race or ethnicity than the target). However, there may be some innate crowdworker biases that affect whether names were changed or not in these cases and also, possibly which names they were changed to. One option to address any possible biases in names that result from crowdsourcing could be to run another post hoc stage of heuristic name perturbation~\citep{smith-etal-2021-hi} to ablate the contribution of demographic information on names altogether. We leave this option for follow up work.  

A final qualification about our annotator pool is that it represents a demographically skewed sample (see \autoref{tab:turkerrace-gender} and \autoref{tab:turkerage}), meaning that annotators may overlook experiences or other considerations that are not in accordance with their lived experiences. We worried that people might be worse at perturbing text when the target demographic attribute mismatched with their own identities, but anecdotally we did not find many problematic examples of this (although a more systematic, quantitative investigation of this would be ideal). While utilizing crowdsourced data avoids many issues that arise from synthetic data (i.e., grammatical issues and unnaturalness), crowdsourcing has its own limitations, such as human error. We carefully considered the decision to crowdsource annotations before embarking on this work.

\paragraph{The Hard Problem of Measuring Fairness in NLP:} We have argued on the basis of several metrics that \fairberta~is generally more fair than the original RoBERTa model. However, this argument has to be tempered with the very real fact that our current fairness metrics are imperfect.  Many of the standard fairness metrics in NLP have flaws \citep{blodgett-etal-2021-stereotyping}, and often different fairness metrics fail to agree \citep{delobelle-etal-2021-measuring, goldfarb-tarrant-etal-2021-intrinsic, cao-etal-2022-intrinsic}. How best to measure fairness in NLP is an ongoing and open research direction that is far from settled. For example, how best to estimate group-wise performance disparities \citep{lum-etal-2022-debiasing} even for the fairly simple case of binary classification is still actively debated. We have made our best attempt here to measure NLP fairness using (i) common metrics whose weaknesses have been cataloged and can be factored into our interpretation of our results, or (ii) metrics that offer a wider holistic perspective on possible LM biases. 

A related issue pertains to whether intrinsic or extrinsic bias measurements are preferable. Intrinsic metrics probe the underlying LM, whereas extrinsic metrics evaluate models on downstream tasks. Intrinsic metrics are useful, since they target the pretrained model, which under the currently dominant pretrain-then-finetune paradigm, forms the basis for the model regardless of the downstream tasks. However, there is always a worry that intrinsic metrics may not be predictive of what happens downstream, as finetuning can overwrite some of what is learned in pretraining \citep{zhao-bethard-2020-berts, he-etal-2021-analyzing}, a situation we call ``catastrophic forgetting'', when something we like gets overwritten \citep{mccloskey-cohen-1989-catastrophic, goodfellow-etal-2013-empirical, chen-etal-2020-recall}.  The empirical results are mixed, with some works finding that debiasing the pretrained model before finetuning does benefit downstream tasks \citep{jin-etal-2021-transferability}, but others find that intrinsic and extrinsic bias measurements do not correlate \citep{goldfarb-tarrant-etal-2021-intrinsic, cao-etal-2022-intrinsic}, raising questions about which approach to trust more. 

For our part, we explore both intrinsic and extrinsic measurements: we use intrinsic measurements to evaluate the fairness of \fairberta~(CrowS-Pairs, WEAT/SEAT, HB), and then explore extrinsic fairness downstream with the \fairscore. We find that debiasing during both pretraining and finetuning stages reduces model bias. 
% We also investigate extrinsic datasets for measuring social bias in question answering \citep{parrish-etal-2021-bbq} and NLI \citep{rudinger2018gender, wang-etal-2019-superglue} and find that extrinsic metric performance is correlated with underlying task performance (see \autoref{sec:extrinsicmetrics}), i.e. a model that is better at question answering tasks is likely to perform better on a question answering based fairness metric, though it may still exhibit biases. 
It is not the goal of this paper to argue in favor of one kind of measurement over the other, and as new metrics and approaches for bias measurement are innovated, we hope to continue to benchmark our \fairberta~model against them. 
 
\paragraph{Recall and Precision in Perturbability Scoring:} It was difficult to select snippets from a large text sample to present to annotators, because there's no perfect way to select only and all snippets with demographic references in them during preprocessing.  We relied upon a terms list from \newcite{ma-etal-2021-dynaboard}, and then asked humans to verify that the snippets indeed do contain references to demographic attributes. One could imagine employing humans to sift through all examples in the pretraining data, looking for perturbable words, but this was prohibitively expensive. In short, our preproccessing optimized for precision over recall, because, in general, the majority of source snippets do not have demographic references, and we wanted to be judicious with annotators' time. This means that it's possible that we overlooked possibly perturbable examples, because they didn't contain terms on the words list we used in preprocessing. Future work could explore better ways of finding perturbable snippets, for example, using another neural model for preprocessing.

\paragraph{Demographic Perturbations in English:} Currently, we have focused solely on English (as spoken by workers on Amazon Mechanical Turk). However, one could imagine extending our dataset and approach to other languages as well. To do this, one could draw inspiration from existing work on computational morphology. A form of gender rewriting, i.e., morphological reinflection for grammatical gender, has already been used to combat bias arising from morphological underspecification during automatic translation for languages with more grammatical gender morphology than English, including Arabic \cite{habash-etal-2019-automatic, alhafni-etal-2020-gender} and Italian \cite{vanmassenhove-monti-2021-gender}. These works differ from our setting in that they narrowly focused on morphological changes related to grammatical gender inflection and are not defined for the other axes (age, race)---even for the gender axis for which the morphological task is defined, in the general case, one wouldn't want to replace whole words, such \textit{son} to \textit{daughter}, but would only change the form of words that share a lemma, as for Italian changing \textit{maestr-\textbf{o}} `teacher.MASC' to \textit{maestr-\textbf{a}} `teacher.FEM'. %This body of work is likely to yield insights for research like ours though, and the idea of 
Extending our perturbation approaches to languages with more extensive morphological gender marking than English would be an interesting avenue for future work.

% \begin{itemize}
%     \item to/as/about axes (only looking at ABOUT)
%     \item equity/equality
% \end{itemize}

% \section*{Acknowledgments}
% Researchers and engineers at FAIR, HuggingFace, NYU etc. who commented on or otherwise supported this work, including Melanie Kambudur, Stephen Roller, Keren Fuentes, Kushal Tirumala, Koustuv Sinha, Naman Goyal, Tristan Thrush, Mona Diab, Chinnadhurai Sankar, Jane Yu, Alon Halevy, Yi-Chia Wang, Olga Golovneva, Armen Aghajanyan and Qinqing Zheng. Thanks also for early feedback on drafts from Yacine Jernite, Hagen Blix, Hannah Rose Kirk, Katerina Margatina, Irene Solaiman, Meg Mitchell, Zeerak Talat, Anne Lauscher, Maarten Sap, Nathan Ng, Bertie Vidgen, Nikita Nangia, Paul R\"{o}ttger and the attendees of the Meta AI Responsible NLP Project Sync. Any errors or omissions are ours, of course!

\bibliographystyle{acl_natbib}
\bibliography{main, anthology}

\begin{thebibliography}{97}
\expandafter\ifx\csname natexlab\endcsname\relax\def\natexlab#1{#1}\fi

\bibitem[{Ackerman(2019)}]{ackerman-2019-syntactic}
Lauren~M Ackerman. 2019.
\newblock \href {https://www.glossa-journal.org/article/id/5224/} {Syntactic
  and cognitive issues in investigating gendered coreference}.
\newblock \emph{Glossa}.

\bibitem[{Alhafni et~al.(2020)Alhafni, Habash, and
  Bouamor}]{alhafni-etal-2020-gender}
Bashar Alhafni, Nizar Habash, and Houda Bouamor. 2020.
\newblock \href {https://aclanthology.org/2020.gebnlp-1.12} {Gender-aware
  reinflection using linguistically enhanced neural models}.
\newblock In \emph{Proceedings of the Second Workshop on Gender Bias in Natural
  Language Processing}, pages 139--150, Barcelona, Spain (Online). Association
  for Computational Linguistics.

\bibitem[{Bailey et~al.(2022)Bailey, Williams, and
  Cimpian}]{bailey-etal-2022-based}
April~H. Bailey, Adina Williams, and Andrei Cimpian. 2022.
\newblock \href {https://www.science.org/doi/10.1126/sciadv.abm2463} {Based on
  billions of words on the internet, people= men}.
\newblock \emph{Science Advances}, 8(13):eabm2463.

\bibitem[{Barocas et~al.(2017)Barocas, Crawford, Shapiro, and
  Wallach}]{barocas-etal-2017-problem}
Solon Barocas, Kate Crawford, Aaron Shapiro, and Hanna Wallach. 2017.
\newblock The problem with bias: from allocative to representational harms in
  machine learning.
\newblock In \emph{{Special Interest Group for Computing, Information and
  Society (SIGCIS)}}, volume~2.

\bibitem[{Bender and Friedman(2018)}]{bender-friedman-2018-data}
Emily~M. Bender and Batya Friedman. 2018.
\newblock \href {https://doi.org/10.1162/tacl_a_00041} {Data statements for
  natural language processing: Toward mitigating system bias and enabling
  better science}.
\newblock \emph{Transactions of the Association for Computational Linguistics},
  6:587--604.

\bibitem[{Bender et~al.(2021)Bender, Gebru, McMillan-Major, and
  Shmitchell}]{bender2021dangers}
Emily~M Bender, Timnit Gebru, Angelina McMillan-Major, and Shmargaret
  Shmitchell. 2021.
\newblock \href {https://dl.acm.org/doi/10.1145/3442188.3445922} {On the
  dangers of stochastic parrots: Can language models be too big?}
\newblock In \emph{Proceedings of the 2021 ACM Conference on Fairness,
  Accountability, and Transparency}, pages 610--623.

\bibitem[{Bentivogli et~al.(2009)Bentivogli, Clark, Dagan, and
  Giampiccolo}]{bentivogli-etal-2009-fifth}
Luisa Bentivogli, Peter Clark, Ido Dagan, and Danilo Giampiccolo. 2009.
\newblock \href
  {https://tac.nist.gov/publications/2009/additional.papers/RTE5_overview.proceedings.pdf}
  {The fifth pascal recognizing textual entailment challenge.}
\newblock In \emph{TAC}.

\bibitem[{Berk et~al.(2017)Berk, Heidari, Jabbari, Joseph, Kearns, Morgenstern,
  Neel, and Roth}]{berk-etal-2017-convex}
Richard Berk, Hoda Heidari, Shahin Jabbari, Matthew Joseph, Michael Kearns,
  Jamie Morgenstern, Seth Neel, and Aaron Roth. 2017.
\newblock \href {https://arxiv.org/abs/1706.02409} {A convex framework for fair
  regression}.
\newblock \emph{arXiv preprint arXiv:1706.02409}.

\bibitem[{Beydoun(2015)}]{beydoun-etal-2015-boxed}
Khaled~A Beydoun. 2015.
\newblock \href {https://papers.ssrn.com/sol3/papers.cfm?abstract_id=2760604}
  {Boxed in: Reclassification of arab americans on the us census as progress or
  peril}.
\newblock \emph{Loy. U. Chi. LJ}, 47:693.

\bibitem[{Blodgett et~al.(2021)Blodgett, Lopez, Olteanu, Sim, and
  Wallach}]{blodgett-etal-2021-stereotyping}
Su~Lin Blodgett, Gilsinia Lopez, Alexandra Olteanu, Robert Sim, and Hanna
  Wallach. 2021.
\newblock \href {https://doi.org/10.18653/v1/2021.acl-long.81} {Stereotyping
  {N}orwegian salmon: An inventory of pitfalls in fairness benchmark datasets}.
\newblock In \emph{Proceedings of the 59th Annual Meeting of the Association
  for Computational Linguistics and the 11th International Joint Conference on
  Natural Language Processing (Volume 1: Long Papers)}, pages 1004--1015,
  Online. Association for Computational Linguistics.

\bibitem[{Buolamwini and Gebru(2018)}]{buolamwini-gebru-2018-gender}
Joy Buolamwini and Timnit Gebru. 2018.
\newblock \href
  {http://proceedings.mlr.press/v81/buolamwini18a/buolamwini18a.pdf} {Gender
  shades: Intersectional accuracy disparities in commercial gender
  classification}.
\newblock In \emph{Conference on fairness, accountability and transparency},
  pages 77--91. PMLR.

\bibitem[{Caliskan et~al.(2017)Caliskan, Bryson, and
  Narayanan}]{caliskan-etal-2017-semantics}
Aylin Caliskan, Joanna~J Bryson, and Arvind Narayanan. 2017.
\newblock \href {https://pubmed.ncbi.nlm.nih.gov/28408601/} {Semantics derived
  automatically from language corpora contain human-like biases}.
\newblock \emph{Science}, 356(6334):183--186.

\bibitem[{Cao et~al.(2022)Cao, Pruksachatkun, Chang, Gupta, Kumar, Dhamala, and
  Galstyan}]{cao-etal-2022-intrinsic}
Yang Cao, Yada Pruksachatkun, Kai-Wei Chang, Rahul Gupta, Varun Kumar, Jwala
  Dhamala, and Aram Galstyan. 2022.
\newblock \href {https://aclanthology.org/2022.acl-short.62} {On the intrinsic
  and extrinsic fairness evaluation metrics for contextualized language
  representations}.
\newblock In \emph{Proceedings of the 60th Annual Meeting of the Association
  for Computational Linguistics (Volume 2: Short Papers)}, pages 561--570,
  Dublin, Ireland. Association for Computational Linguistics.

\bibitem[{Cer et~al.(2017)Cer, Diab, Agirre, Lopez-Gazpio, and
  Specia}]{cer-etal-2017-semeval}
Daniel Cer, Mona Diab, Eneko Agirre, I{\~n}igo Lopez-Gazpio, and Lucia Specia.
  2017.
\newblock \href {https://doi.org/10.18653/v1/S17-2001} {{S}em{E}val-2017 task
  1: Semantic textual similarity multilingual and crosslingual focused
  evaluation}.
\newblock In \emph{Proceedings of the 11th International Workshop on Semantic
  Evaluation ({S}em{E}val-2017)}, pages 1--14, Vancouver, Canada. Association
  for Computational Linguistics.

\bibitem[{Chang et~al.(2019)Chang, Prabhakaran, and
  Ordonez}]{chang-etal-2019-bias}
Kai-Wei Chang, Vinodkumar Prabhakaran, and Vicente Ordonez. 2019.
\newblock \href {https://aclanthology.org/D19-2004} {Bias and fairness in
  natural language processing}.
\newblock In \emph{Proceedings of the 2019 Conference on Empirical Methods in
  Natural Language Processing and the 9th International Joint Conference on
  Natural Language Processing (EMNLP-IJCNLP): Tutorial Abstracts}, Hong Kong,
  China. Association for Computational Linguistics.

\bibitem[{Chen et~al.(2020)Chen, Hou, Cui, Che, Liu, and
  Yu}]{chen-etal-2020-recall}
Sanyuan Chen, Yutai Hou, Yiming Cui, Wanxiang Che, Ting Liu, and Xiangzhan Yu.
  2020.
\newblock \href {https://doi.org/10.18653/v1/2020.emnlp-main.634} {Recall and
  learn: Fine-tuning deep pretrained language models with less forgetting}.
\newblock In \emph{Proceedings of the 2020 Conference on Empirical Methods in
  Natural Language Processing (EMNLP)}, pages 7870--7881, Online. Association
  for Computational Linguistics.

\bibitem[{Chowdhery et~al.(2022)Chowdhery, Narang, Devlin, Bosma, Mishra,
  Roberts, Barham, Chung, Sutton, Gehrmann et~al.}]{chowdhery-etal-2022-palm}
Aakanksha Chowdhery, Sharan Narang, Jacob Devlin, Maarten Bosma, Gaurav Mishra,
  Adam Roberts, Paul Barham, Hyung~Won Chung, Charles Sutton, Sebastian
  Gehrmann, et~al. 2022.
\newblock \href {https://arxiv.org/abs/2204.02311} {Pa{LM}: Scaling language
  modeling with pathways}.
\newblock \emph{arXiv preprint arXiv:2204.02311}.

\bibitem[{Dagan et~al.(2005)Dagan, Glickman, and
  Magnini}]{dagan-etal-2005-pascal}
Ido Dagan, Oren Glickman, and Bernardo Magnini. 2005.
\newblock \href
  {https://tac.nist.gov/publications/2009/additional.papers/RTE5_overview.proceedings.pdf}
  {The pascal recognising textual entailment challenge}.
\newblock In \emph{Machine Learning Challenges Workshop}, pages 177--190.
  Springer.

\bibitem[{De-Arteaga et~al.(2019)De-Arteaga, Romanov, Wallach, Chayes, Borgs,
  Chouldechova, Geyik, Kenthapadi, and Kalai}]{DeArteaga2019BiasIB}
Maria De-Arteaga, Alexey Romanov, Hanna~M. Wallach, Jennifer~T. Chayes,
  Christian Borgs, Alexandra Chouldechova, Sahin~Cem Geyik, Krishnaram
  Kenthapadi, and Adam~Tauman Kalai. 2019.
\newblock \href {https://dl.acm.org/doi/10.1145/3287560.3287572} {Bias in bios:
  A case study of semantic representation bias in a high-stakes setting}.
\newblock \emph{Proceedings of the Conference on Fairness, Accountability, and
  Transparency}.

\bibitem[{Delobelle et~al.(2021)Delobelle, Tokpo, Calders, and
  Berendt}]{delobelle-etal-2021-measuring}
Pieter Delobelle, Ewoenam~Kwaku Tokpo, Toon Calders, and Bettina Berendt. 2021.
\newblock \href {https://arxiv.org/abs/2112.07447v1} {Measuring fairness with
  biased rulers: A survey on quantifying biases in pretrained language models}.
\newblock \emph{arXiv preprint arXiv:2112.07447}.

\bibitem[{Dinan et~al.(2020{\natexlab{a}})Dinan, Fan, Williams, Urbanek, Kiela,
  and Weston}]{dinan-etal-2020-queens}
Emily Dinan, Angela Fan, Adina Williams, Jack Urbanek, Douwe Kiela, and Jason
  Weston. 2020{\natexlab{a}}.
\newblock \href {https://doi.org/10.18653/v1/2020.emnlp-main.656} {Queens are
  powerful too: Mitigating gender bias in dialogue generation}.
\newblock In \emph{Proceedings of the 2020 Conference on Empirical Methods in
  Natural Language Processing (EMNLP)}, pages 8173--8188, Online. Association
  for Computational Linguistics.

\bibitem[{Dinan et~al.(2020{\natexlab{b}})Dinan, Fan, Wu, Weston, Kiela, and
  Williams}]{dinan-etal-2020-multi}
Emily Dinan, Angela Fan, Ledell Wu, Jason Weston, Douwe Kiela, and Adina
  Williams. 2020{\natexlab{b}}.
\newblock \href {https://doi.org/10.18653/v1/2020.emnlp-main.23}
  {Multi-dimensional gender bias classification}.
\newblock In \emph{Proceedings of the 2020 Conference on Empirical Methods in
  Natural Language Processing (EMNLP)}, pages 314--331, Online. Association for
  Computational Linguistics.

\bibitem[{Dinan et~al.(2019)Dinan, Humeau, Chintagunta, and
  Weston}]{dinan-etal-2019-build}
Emily Dinan, Samuel Humeau, Bharath Chintagunta, and Jason Weston. 2019.
\newblock \href {https://doi.org/10.18653/v1/D19-1461} {Build it break it fix
  it for dialogue safety: Robustness from adversarial human attack}.
\newblock In \emph{Proceedings of the 2019 Conference on Empirical Methods in
  Natural Language Processing and the 9th International Joint Conference on
  Natural Language Processing (EMNLP-IJCNLP)}, pages 4537--4546, Hong Kong,
  China. Association for Computational Linguistics.

\bibitem[{Dua et~al.(2019)Dua, Wang, Dasigi, Stanovsky, Singh, and
  Gardner}]{dua-etal-2019-drop}
Dheeru Dua, Yizhong Wang, Pradeep Dasigi, Gabriel Stanovsky, Sameer Singh, and
  Matt Gardner. 2019.
\newblock \href {https://doi.org/10.18653/v1/N19-1246} {{DROP}: A reading
  comprehension benchmark requiring discrete reasoning over paragraphs}.
\newblock In \emph{Proceedings of the 2019 Conference of the North {A}merican
  Chapter of the Association for Computational Linguistics: Human Language
  Technologies, Volume 1 (Long and Short Papers)}, pages 2368--2378,
  Minneapolis, Minnesota. Association for Computational Linguistics.

\bibitem[{Emmery et~al.(2022)Emmery, K{\'a}d{\'a}r, Chrupa{\l}a, and
  Daelemans}]{emmery-etal-2022-cyberbullying}
Chris Emmery, {\'A}kos K{\'a}d{\'a}r, Grzegorz Chrupa{\l}a, and Walter
  Daelemans. 2022.
\newblock \href {https://arxiv.org/abs/2201.06384v1} {Cyberbullying classifiers
  are sensitive to model-agnostic perturbations}.
\newblock \emph{arXiv preprint arXiv:2201.06384}.

\bibitem[{Fan et~al.(2019)Fan, White, Sharma, Su, Choubey, Huang, and
  Wang}]{fan2019plainsight}
Lisa Fan, Marshall White, Eva Sharma, Ruisi Su, Prafulla~Kumar Choubey, Ruihong
  Huang, and Lu~Wang. 2019.
\newblock \href {https://doi.org/10.18653/v1/D19-1664} {In plain sight: Media
  bias through the lens of factual reporting}.
\newblock In \emph{Proceedings of the 2019 Conference on Empirical Methods in
  Natural Language Processing and the 9th International Joint Conference on
  Natural Language Processing (EMNLP-IJCNLP)}, pages 6342--6348, Hong Kong,
  China. Association for Computational Linguistics.

\bibitem[{Field et~al.(2021)Field, Blodgett, Waseem, and
  Tsvetkov}]{field-etal-2021-survey}
Anjalie Field, Su~Lin Blodgett, Zeerak Waseem, and Yulia Tsvetkov. 2021.
\newblock \href {https://doi.org/10.18653/v1/2021.acl-long.149} {A survey of
  race, racism, and anti-racism in {NLP}}.
\newblock In \emph{Proceedings of the 59th Annual Meeting of the Association
  for Computational Linguistics and the 11th International Joint Conference on
  Natural Language Processing (Volume 1: Long Papers)}, pages 1905--1925,
  Online. Association for Computational Linguistics.

\bibitem[{Galinsky et~al.(2003)Galinsky, Hugenberg, Groom, and
  Bodenhausen}]{galinsky-etal-2003-reappropriation}
Adam~D Galinsky, Kurt Hugenberg, Carla Groom, and Galen~V Bodenhausen. 2003.
\newblock \href
  {https://www.emerald.com/insight/content/doi/10.1016/S1534-0856(02)05009-0/full/html}
  {The reappropriation of stigmatizing labels: Implications for social
  identity}.
\newblock In \emph{Identity issues in groups}. Emerald Group Publishing
  Limited.

\bibitem[{Geirhos et~al.(2020)Geirhos, Jacobsen, Michaelis, Zemel, Brendel,
  Bethge, and Wichmann}]{geirhos-etal-2020-shortcut}
Robert Geirhos, J{\"o}rn-Henrik Jacobsen, Claudio Michaelis, Richard Zemel,
  Wieland Brendel, Matthias Bethge, and Felix~A Wichmann. 2020.
\newblock \href {https://www.nature.com/articles/s42256-020-00257-z} {Shortcut
  learning in deep neural networks}.
\newblock \emph{Nature Machine Intelligence}, 2(11):665--673.

\bibitem[{Geva et~al.(2020)Geva, Gupta, and Berant}]{geva-etal-2020-injecting}
Mor Geva, Ankit Gupta, and Jonathan Berant. 2020.
\newblock \href {https://doi.org/10.18653/v1/2020.acl-main.89} {Injecting
  numerical reasoning skills into language models}.
\newblock In \emph{Proceedings of the 58th Annual Meeting of the Association
  for Computational Linguistics}, pages 946--958, Online. Association for
  Computational Linguistics.

\bibitem[{Giampiccolo et~al.(2007)Giampiccolo, Magnini, Dagan, and
  Dolan}]{giampiccolo-etal-2007-third}
Danilo Giampiccolo, Bernardo Magnini, Ido Dagan, and Bill Dolan. 2007.
\newblock \href {https://aclanthology.org/W07-1401} {The third {PASCAL}
  recognizing textual entailment challenge}.
\newblock In \emph{Proceedings of the {ACL}-{PASCAL} Workshop on Textual
  Entailment and Paraphrasing}, pages 1--9, Prague. Association for
  Computational Linguistics.

\bibitem[{Goldfarb-Tarrant et~al.(2021)Goldfarb-Tarrant, Marchant,
  Mu{\~n}oz~S{\'a}nchez, Pandya, and
  Lopez}]{goldfarb-tarrant-etal-2021-intrinsic}
Seraphina Goldfarb-Tarrant, Rebecca Marchant, Ricardo Mu{\~n}oz~S{\'a}nchez,
  Mugdha Pandya, and Adam Lopez. 2021.
\newblock \href {https://doi.org/10.18653/v1/2021.acl-long.150} {Intrinsic bias
  metrics do not correlate with application bias}.
\newblock In \emph{Proceedings of the 59th Annual Meeting of the Association
  for Computational Linguistics and the 11th International Joint Conference on
  Natural Language Processing (Volume 1: Long Papers)}, pages 1926--1940,
  Online. Association for Computational Linguistics.

\bibitem[{Goodfellow et~al.(2013)Goodfellow, Mirza, Xiao, Courville, and
  Bengio}]{goodfellow-etal-2013-empirical}
Ian~J Goodfellow, Mehdi Mirza, Da~Xiao, Aaron Courville, and Yoshua Bengio.
  2013.
\newblock \href {https://arxiv.org/abs/1312.6211} {An empirical investigation
  of catastrophic forgetting in gradient-based neural networks}.
\newblock \emph{arXiv preprint arXiv:1312.6211}.

\bibitem[{Habash et~al.(2019)Habash, Bouamor, and
  Chung}]{habash-etal-2019-automatic}
Nizar Habash, Houda Bouamor, and Christine Chung. 2019.
\newblock \href {https://doi.org/10.18653/v1/W19-3822} {Automatic gender
  identification and reinflection in {A}rabic}.
\newblock In \emph{Proceedings of the First Workshop on Gender Bias in Natural
  Language Processing}, pages 155--165, Florence, Italy. Association for
  Computational Linguistics.

\bibitem[{Haim et~al.(2006)Haim, Dagan, Dolan, Ferro, Giampiccolo, Magnini, and
  Szpektor}]{haim-etal-2006-second}
R~Bar Haim, Ido Dagan, Bill Dolan, Lisa Ferro, Danilo Giampiccolo, Bernardo
  Magnini, and Idan Szpektor. 2006.
\newblock \href {https://roybarhaim.webs.com/Papers/RTE2-organizers.pdf} {The
  second pascal recognising textual entailment challenge}.
\newblock In \emph{Proceedings of the Second PASCAL Challenges Workshop on
  Recognising Textual Entailment}, volume~7.

\bibitem[{Hall~Maudslay et~al.(2019)Hall~Maudslay, Gonen, Cotterell, and
  Teufel}]{hall-maudslay-etal-2019-name}
Rowan Hall~Maudslay, Hila Gonen, Ryan Cotterell, and Simone Teufel. 2019.
\newblock \href {https://doi.org/10.18653/v1/D19-1530} {It{'}s all in the name:
  Mitigating gender bias with name-based counterfactual data substitution}.
\newblock In \emph{Proceedings of the 2019 Conference on Empirical Methods in
  Natural Language Processing and the 9th International Joint Conference on
  Natural Language Processing (EMNLP-IJCNLP)}, pages 5267--5275, Hong Kong,
  China. Association for Computational Linguistics.

\bibitem[{Haller et~al.(2006)Haller, Dorries, and
  Rahn}]{haller-etal-2006-media}
Beth Haller, Bruce Dorries, and Jessica Rahn. 2006.
\newblock \href
  {https://www.tandfonline.com/doi/abs/10.1080/09687590500375416?journalCode=cdso20}
  {Media labeling versus the us disability community identity: a study of
  shifting cultural language}.
\newblock \emph{Disability \& Society}, 21(1):61--75.

\bibitem[{He et~al.(2021)He, Liu, Cho, Ott, Liu, Glass, and
  Peng}]{he-etal-2021-analyzing}
Tianxing He, Jun Liu, Kyunghyun Cho, Myle Ott, Bing Liu, James Glass, and
  Fuchun Peng. 2021.
\newblock \href {https://doi.org/10.18653/v1/2021.eacl-main.95} {Analyzing the
  forgetting problem in pretrain-finetuning of open-domain dialogue response
  models}.
\newblock In \emph{Proceedings of the 16th Conference of the European Chapter
  of the Association for Computational Linguistics: Main Volume}, pages
  1121--1133, Online. Association for Computational Linguistics.

\bibitem[{Hendricks et~al.(2018)Hendricks, Burns, Saenko, Darrell, and
  Rohrbach}]{hendricks-etal-2018-women}
Lisa~Anne Hendricks, Kaylee Burns, Kate Saenko, Trevor Darrell, and Anna
  Rohrbach. 2018.
\newblock \href {https://arxiv.org/abs/1803.09797} {Women also snowboard:
  Overcoming bias in captioning models}.
\newblock In \emph{Proceedings of the European Conference on Computer Vision
  (ECCV)}, pages 771--787.

\bibitem[{Jin et~al.(2021)Jin, Barbieri, Kennedy, Mostafazadeh~Davani, Neves,
  and Ren}]{jin-etal-2021-transferability}
Xisen Jin, Francesco Barbieri, Brendan Kennedy, Aida Mostafazadeh~Davani,
  Leonardo Neves, and Xiang Ren. 2021.
\newblock \href {https://doi.org/10.18653/v1/2021.naacl-main.296} {On
  transferability of bias mitigation effects in language model fine-tuning}.
\newblock In \emph{Proceedings of the 2021 Conference of the North American
  Chapter of the Association for Computational Linguistics: Human Language
  Technologies}, pages 3770--3783, Online. Association for Computational
  Linguistics.

\bibitem[{Kaplan et~al.(2020)Kaplan, McCandlish, Henighan, Brown, Chess, Child,
  Gray, Radford, Wu, and Amodei}]{kaplan-etal-2020-scaling}
Jared Kaplan, Sam McCandlish, Tom Henighan, Tom~B. Brown, Benjamin Chess, Rewon
  Child, Scott Gray, Alec Radford, Jeffrey Wu, and Dario Amodei. 2020.
\newblock \href {http://arxiv.org/abs/2001.08361} {Scaling laws for neural
  language models}.
\newblock \emph{arxiv}, abs/2001.08361.

\bibitem[{Kayyali(2013)}]{kayyali-etal-2013-us}
Randa Kayyali. 2013.
\newblock \href
  {https://www.tandfonline.com/doi/abs/10.1080/1369183X.2013.778150} {Us census
  classifications and arab americans: contestations and definitions of identity
  markers}.
\newblock \emph{Journal of Ethnic and Migration Studies}, 39(8):1299--1318.

\bibitem[{Keyes(2019)}]{keyes-2019-counting}
Os~Keyes. 2019.
\newblock \href {https://reallifemag.com/counting-the-countless/} {Counting the
  countless: Why data science is a profound threat for queer people}.
\newblock \emph{Real Life}, 2.

\bibitem[{Lai et~al.(2017)Lai, Xie, Liu, Yang, and Hovy}]{lai-etal-2017-race}
Guokun Lai, Qizhe Xie, Hanxiao Liu, Yiming Yang, and Eduard Hovy. 2017.
\newblock \href {https://doi.org/10.18653/v1/D17-1082} {{RACE}: Large-scale
  {R}e{A}ding comprehension dataset from examinations}.
\newblock In \emph{Proceedings of the 2017 Conference on Empirical Methods in
  Natural Language Processing}, pages 785--794, Copenhagen, Denmark.
  Association for Computational Linguistics.

\bibitem[{Lauscher et~al.(2022)Lauscher, Crowley, and
  Hovy}]{lauscher-etal-2022-welcome}
Anne Lauscher, Archie Crowley, and Dirk Hovy. 2022.
\newblock \href {https://arxiv.org/abs/2202.11923} {Welcome to the modern world
  of pronouns: Identity-inclusive natural language processing beyond gender}.
\newblock \emph{arXiv preprint arXiv:2202.11923}.

\bibitem[{Lewis et~al.(2020)Lewis, Liu, Goyal, Ghazvininejad, Mohamed, Levy,
  Stoyanov, and Zettlemoyer}]{lewis-etal-2020-bart}
Mike Lewis, Yinhan Liu, Naman Goyal, Marjan Ghazvininejad, Abdelrahman Mohamed,
  Omer Levy, Veselin Stoyanov, and Luke Zettlemoyer. 2020.
\newblock \href {https://doi.org/10.18653/v1/2020.acl-main.703} {{BART}:
  Denoising sequence-to-sequence pre-training for natural language generation,
  translation, and comprehension}.
\newblock In \emph{Proceedings of the 58th Annual Meeting of the Association
  for Computational Linguistics}, pages 7871--7880, Online. Association for
  Computational Linguistics.

\bibitem[{Lin et~al.(2020)Lin, Lee, Khanna, and Ren}]{lin-etal-2020-birds}
Bill~Yuchen Lin, Seyeon Lee, Rahul Khanna, and Xiang Ren. 2020.
\newblock \href {https://doi.org/10.18653/v1/2020.emnlp-main.557} {{B}irds have
  four legs?! {N}umer{S}ense: {P}robing {N}umerical {C}ommonsense {K}nowledge
  of {P}re-{T}rained {L}anguage {M}odels}.
\newblock In \emph{Proceedings of the 2020 Conference on Empirical Methods in
  Natural Language Processing (EMNLP)}, pages 6862--6868, Online. Association
  for Computational Linguistics.

\bibitem[{Liu et~al.(2019)Liu, Ott, Goyal, Du, Joshi, Chen, Levy, Lewis,
  Zettlemoyer, and Stoyanov}]{liu-etal-2019-roberta}
Yinhan Liu, Myle Ott, Naman Goyal, Jingfei Du, Mandar Joshi, Danqi Chen, Omer
  Levy, Mike Lewis, Luke Zettlemoyer, and Veselin Stoyanov. 2019.
\newblock \href {https://arxiv.org/abs/1907.11692} {Ro{BERT}a: A robustly
  optimized {BERT} pretraining approach}.
\newblock \emph{arXiv preprint arXiv:1907.11692}.

\bibitem[{Lum et~al.(2022)Lum, Zhang, and Bower}]{lum-etal-2022-debiasing}
Kristian Lum, Yunfeng Zhang, and Amanda Bower. 2022.
\newblock \href {https://arxiv.org/abs/2205.05770} {De-biasing "bias"
  measurement}.
\newblock In \emph{Proceedings of the 2022 ACM Conference on Fairness,
  Accountability, and Transparency}, FAccT '22, New York, NY, USA. Association
  for Computing Machinery.

\bibitem[{Ma et~al.(2021)Ma, Ethayarajh, Thrush, Jain, Wu, Jia, Potts,
  Williams, and Kiela}]{ma-etal-2021-dynaboard}
Zhiyi Ma, Kawin Ethayarajh, Tristan Thrush, Somya Jain, Ledell Wu, Robin Jia,
  Christopher Potts, Adina Williams, and Douwe Kiela. 2021.
\newblock \href
  {https://papers.nips.cc/paper/2021/hash/55b1927fdafef39c48e5b73b5d61ea60-Abstract.html}
  {Dynaboard: An evaluation-as-a-service platform for holistic next-generation
  benchmarking}.
\newblock \emph{Advances in Neural Information Processing Systems}, 34.

\bibitem[{May et~al.(2019)May, Wang, Bordia, Bowman, and
  Rudinger}]{may-etal-2019-measuring}
Chandler May, Alex Wang, Shikha Bordia, Samuel~R. Bowman, and Rachel Rudinger.
  2019.
\newblock \href {https://doi.org/10.18653/v1/N19-1063} {On measuring social
  biases in sentence encoders}.
\newblock In \emph{Proceedings of the 2019 Conference of the North {A}merican
  Chapter of the Association for Computational Linguistics: Human Language
  Technologies, Volume 1 (Long and Short Papers)}, pages 622--628, Minneapolis,
  Minnesota. Association for Computational Linguistics.

\bibitem[{McCloskey and Cohen(1989)}]{mccloskey-cohen-1989-catastrophic}
Michael McCloskey and Neal~J Cohen. 1989.
\newblock Catastrophic interference in connectionist networks: The sequential
  learning problem.
\newblock In \emph{Psychology of learning and motivation}, volume~24, pages
  109--165. Elsevier.

\bibitem[{Mehrabi et~al.(2021)Mehrabi, Zhou, Morstatter, Pujara, Ren, and
  Galstyan}]{mehrabi-etal-2021-lawyers}
Ninareh Mehrabi, Pei Zhou, Fred Morstatter, Jay Pujara, Xiang Ren, and Aram
  Galstyan. 2021.
\newblock \href {https://doi.org/10.18653/v1/2021.emnlp-main.410} {Lawyers are
  dishonest? quantifying representational harms in commonsense knowledge
  resources}.
\newblock In \emph{Proceedings of the 2021 Conference on Empirical Methods in
  Natural Language Processing}, pages 5016--5033, Online and Punta Cana,
  Dominican Republic. Association for Computational Linguistics.

\bibitem[{Menon and Williamson(2018)}]{pmlr-etal-menon-2018}
Aditya~Krishna Menon and Robert~C Williamson. 2018.
\newblock \href {https://proceedings.mlr.press/v81/menon18a.html} {The cost of
  fairness in binary classification}.
\newblock In \emph{Proceedings of the 1st Conference on Fairness,
  Accountability and Transparency}, volume~81 of \emph{Proceedings of Machine
  Learning Research}, pages 107--118. PMLR.

\bibitem[{Merullo et~al.(2019)Merullo, Yeh, Handler, Grissom~II, O{'}Connor,
  and Iyyer}]{merullo-etal-2019}
Jack Merullo, Luke Yeh, Abram Handler, Alvin Grissom~II, Brendan O{'}Connor,
  and Mohit Iyyer. 2019.
\newblock \href {https://doi.org/10.18653/v1/D19-1666} {Investigating sports
  commentator bias within a large corpus of {A}merican football broadcasts}.
\newblock In \emph{Proceedings of the 2019 Conference on Empirical Methods in
  Natural Language Processing and the 9th International Joint Conference on
  Natural Language Processing (EMNLP-IJCNLP)}, pages 6354--6360, Hong Kong,
  China. Association for Computational Linguistics.

\bibitem[{Miller et~al.(2017)Miller, Feng, Batra, Bordes, Fisch, Lu, Parikh,
  and Weston}]{miller-etal-2017-parlai}
Alexander Miller, Will Feng, Dhruv Batra, Antoine Bordes, Adam Fisch, Jiasen
  Lu, Devi Parikh, and Jason Weston. 2017.
\newblock \href {https://doi.org/10.18653/v1/D17-2014} {{P}arl{AI}: A dialog
  research software platform}.
\newblock In \emph{Proceedings of the 2017 Conference on Empirical Methods in
  Natural Language Processing: System Demonstrations}, pages 79--84,
  Copenhagen, Denmark. Association for Computational Linguistics.

\bibitem[{Moss et~al.(2020)Moss, Rosenzweig, Robinson, and
  Litman}]{moss-etal-2020-demographic}
Aaron~J Moss, Cheskie Rosenzweig, Jonathan Robinson, and Leib Litman. 2020.
\newblock \href
  {https://www.cell.com/trends/cognitive-sciences/fulltext/S1364-6613(20)30138-8}
  {Demographic stability on mechanical turk despite covid-19}.
\newblock \emph{Trends in cognitive sciences}, 24(9):678--680.

\bibitem[{Nangia et~al.(2020)Nangia, Vania, Bhalerao, and
  Bowman}]{nangia-etal-2020-crows}
Nikita Nangia, Clara Vania, Rasika Bhalerao, and Samuel~R. Bowman. 2020.
\newblock \href {https://doi.org/10.18653/v1/2020.emnlp-main.154}
  {{C}row{S}-pairs: A challenge dataset for measuring social biases in masked
  language models}.
\newblock In \emph{Proceedings of the 2020 Conference on Empirical Methods in
  Natural Language Processing (EMNLP)}, pages 1953--1967, Online. Association
  for Computational Linguistics.

\bibitem[{Nie et~al.(2020)Nie, Williams, Dinan, Bansal, Weston, and
  Kiela}]{nie-etal-2020-adversarial}
Yixin Nie, Adina Williams, Emily Dinan, Mohit Bansal, Jason Weston, and Douwe
  Kiela. 2020.
\newblock \href {https://doi.org/10.18653/v1/2020.acl-main.441} {Adversarial
  {NLI}: A new benchmark for natural language understanding}.
\newblock In \emph{Proceedings of the 58th Annual Meeting of the Association
  for Computational Linguistics}, pages 4885--4901, Online. Association for
  Computational Linguistics.

\bibitem[{Ott et~al.(2019)Ott, Edunov, Baevski, Fan, Gross, Ng, Grangier, and
  Auli}]{ott2019fairseq}
Myle Ott, Sergey Edunov, Alexei Baevski, Angela Fan, Sam Gross, Nathan Ng,
  David Grangier, and Michael Auli. 2019.
\newblock \href {https://aclanthology.org/N19-4009/} {fairseq: A fast,
  extensible toolkit for sequence modeling}.
\newblock In \emph{Proceedings of NAACL-HLT 2019: Demonstrations}.

\bibitem[{Papakipos and Bitton(2022)}]{papakipos-etal-2022-augly}
Zoe Papakipos and Joanna Bitton. 2022.
\newblock \href {https://arxiv.org/abs/2201.06494} {Augly: Data augmentations
  for robustness}.
\newblock \emph{arXiv preprint arXiv:2201.06494}.

\bibitem[{Papineni et~al.(2002)Papineni, Roukos, Ward, and
  Zhu}]{papineni-etal-2002-bleu}
Kishore Papineni, Salim Roukos, Todd Ward, and Wei-Jing Zhu. 2002.
\newblock \href {https://doi.org/10.3115/1073083.1073135} {{B}leu: a method for
  automatic evaluation of machine translation}.
\newblock In \emph{Proceedings of the 40th Annual Meeting of the Association
  for Computational Linguistics}, pages 311--318, Philadelphia, Pennsylvania,
  USA. Association for Computational Linguistics.

\bibitem[{Parrish et~al.(2021)Parrish, Chen, Nangia, Padmakumar, Phang,
  Thompson, Htut, and Bowman}]{parrish-etal-2021-bbq}
Alicia Parrish, Angelica Chen, Nikita Nangia, Vishakh Padmakumar, Jason Phang,
  Jana Thompson, Phu~Mon Htut, and Samuel~R Bowman. 2021.
\newblock \href {https://arxiv.org/abs/2110.08193} {Bbq: A hand-built bias
  benchmark for question answering}.
\newblock \emph{arXiv preprint arXiv:2110.08193}.

\bibitem[{Prabhakaran et~al.(2019)Prabhakaran, Hutchinson, and
  Mitchell}]{prabhakaran-etal-2019-perturbation}
Vinodkumar Prabhakaran, Ben Hutchinson, and Margaret Mitchell. 2019.
\newblock \href {https://doi.org/10.18653/v1/D19-1578} {Perturbation
  sensitivity analysis to detect unintended model biases}.
\newblock In \emph{Proceedings of the 2019 Conference on Empirical Methods in
  Natural Language Processing and the 9th International Joint Conference on
  Natural Language Processing (EMNLP-IJCNLP)}, pages 5740--5745, Hong Kong,
  China. Association for Computational Linguistics.

\bibitem[{Prates et~al.(2020)Prates, Avelar, and
  Lamb}]{prates-etal-2020-assessing}
Marcelo~OR Prates, Pedro~H Avelar, and Lu{\'\i}s~C Lamb. 2020.
\newblock \href {https://link.springer.com/article/10.1007/s00521-019-04144-6}
  {Assessing gender bias in machine translation: a case study with google
  translate}.
\newblock \emph{Neural Computing and Applications}, 32(10):6363--6381.

\bibitem[{Qi et~al.(2020)Qi, Zhang, Zhang, Bolton, and
  Manning}]{qi-etal-2020-stanza}
Peng Qi, Yuhao Zhang, Yuhui Zhang, Jason Bolton, and Christopher~D. Manning.
  2020.
\newblock \href {https://doi.org/10.18653/v1/2020.acl-demos.14} {{S}tanza: A
  python natural language processing toolkit for many human languages}.
\newblock In \emph{Proceedings of the 58th Annual Meeting of the Association
  for Computational Linguistics: System Demonstrations}, pages 101--108,
  Online. Association for Computational Linguistics.

\bibitem[{Raffel et~al.(2020)Raffel, Shazeer, Roberts, Lee, Narang, Matena,
  Zhou, Li, and Liu}]{raffel-etal-2020-exploring}
Colin Raffel, Noam Shazeer, Adam Roberts, Katherine Lee, Sharan Narang, Michael
  Matena, Yanqi Zhou, Wei Li, and Peter~J. Liu. 2020.
\newblock \href {http://jmlr.org/papers/v21/20-074.html} {Exploring the limits
  of transfer learning with a unified text-to-text transformer}.
\newblock \emph{Journal of Machine Learning Research}, 21:140:1--140:67.

\bibitem[{Rajpurkar et~al.(2016)Rajpurkar, Zhang, Lopyrev, and
  Liang}]{rajpurkar-etal-2016-squad}
Pranav Rajpurkar, Jian Zhang, Konstantin Lopyrev, and Percy Liang. 2016.
\newblock \href {https://doi.org/10.18653/v1/D16-1264} {{SQ}u{AD}: 100,000+
  questions for machine comprehension of text}.
\newblock In \emph{Proceedings of the 2016 Conference on Empirical Methods in
  Natural Language Processing}, pages 2383--2392, Austin, Texas. Association
  for Computational Linguistics.

\bibitem[{Renduchintala and
  Williams(2022)}]{renduchintala-williams-2022-investigating}
Adi Renduchintala and Adina Williams. 2022.
\newblock \href {https://aclanthology.org/2022.acl-long.243} {Investigating
  failures of automatic translationin the case of unambiguous gender}.
\newblock In \emph{Proceedings of the 60th Annual Meeting of the Association
  for Computational Linguistics (Volume 1: Long Papers)}, pages 3454--3469,
  Dublin, Ireland. Association for Computational Linguistics.

\bibitem[{R{\"o}ttger et~al.(2021)R{\"o}ttger, Vidgen, Nguyen, Waseem,
  Margetts, and Pierrehumbert}]{rottger-etal-2021-hatecheck}
Paul R{\"o}ttger, Bertie Vidgen, Dong Nguyen, Zeerak Waseem, Helen Margetts,
  and Janet Pierrehumbert. 2021.
\newblock \href {https://doi.org/10.18653/v1/2021.acl-long.4} {{H}ate{C}heck:
  Functional tests for hate speech detection models}.
\newblock In \emph{Proceedings of the 59th Annual Meeting of the Association
  for Computational Linguistics and the 11th International Joint Conference on
  Natural Language Processing (Volume 1: Long Papers)}, pages 41--58, Online.
  Association for Computational Linguistics.

\bibitem[{Rudinger et~al.(2018)Rudinger, Naradowsky, Leonard, and
  Van~Durme}]{rudinger2018gender}
Rachel Rudinger, Jason Naradowsky, Brian Leonard, and Benjamin Van~Durme. 2018.
\newblock \href {https://aclanthology.org/N18-2002/} {Gender bias in
  coreference resolution}.
\newblock In \emph{Proceedings of the 2018 Conference of the North American
  Chapter of the Association for Computational Linguistics: Human Language
  Technologies, Volume 2 (Short Papers)}, pages 8--14.

\bibitem[{Sedoc and Ungar(2019)}]{sedoc-ungar-2019-role}
Jo{\~a}o Sedoc and Lyle Ungar. 2019.
\newblock \href {https://doi.org/10.18653/v1/W19-3808} {The role of protected
  class word lists in bias identification of contextualized word
  representations}.
\newblock In \emph{Proceedings of the First Workshop on Gender Bias in Natural
  Language Processing}, pages 55--61, Florence, Italy. Association for
  Computational Linguistics.

\bibitem[{Sen et~al.(2022)Sen, Samory, Wagner, and
  Augenstein}]{sen-etal-2022-counterfactually}
Indira Sen, Mattia Samory, Claudia Wagner, and Isabelle Augenstein. 2022.
\newblock \href {https://arxiv.org/abs/2205.04238} {Counterfactually augmented
  data and unintended bias: The case of sexism and hate speech detection}.
\newblock \emph{arXiv preprint arXiv:2205.04238}.

\bibitem[{Smith et~al.(2022)Smith, Hall, Kambadur, Presani, and
  Williams}]{smith-etal-2022-im}
Eric~Michael Smith, Melissa Hall, Melanie Kambadur, Eleonora Presani, and Adina
  Williams. 2022.
\newblock \href {http://arxiv.org/abs/2205.09209} {``{I}'m sorry to hear
  that'': finding bias in language models with a holistic descriptor dataset}.
\newblock \emph{arXiv preprint arXiv:2205.09209}.

\bibitem[{Smith and Williams(2021)}]{smith-etal-2021-hi}
Eric~Michael Smith and Adina Williams. 2021.
\newblock \href {https://arxiv.org/abs/2109.03300} {Hi, my name is {M}artha:
  Using names to measure and mitigate bias in generative dialogue models}.
\newblock \emph{arXiv preprint arXiv:2109.03300}.

\bibitem[{Smith(1992)}]{smith-etal-1992-changing}
Tom~W Smith. 1992.
\newblock \href {https://www.jstor.org/stable/2749204} {Changing racial labels:
  From “colored” to “negro” to “black” to “african american”}.
\newblock \emph{Public Opinion Quarterly}, 56(4):496--514.

\bibitem[{Socher et~al.(2013)Socher, Perelygin, Wu, Chuang, Manning, Ng, and
  Potts}]{socher-etal-2013-recursive}
Richard Socher, Alex Perelygin, Jean Wu, Jason Chuang, Christopher~D. Manning,
  Andrew Ng, and Christopher Potts. 2013.
\newblock \href {https://aclanthology.org/D13-1170} {Recursive deep models for
  semantic compositionality over a sentiment treebank}.
\newblock In \emph{Proceedings of the 2013 Conference on Empirical Methods in
  Natural Language Processing}, pages 1631--1642, Seattle, Washington, USA.
  Association for Computational Linguistics.

\bibitem[{Stock and Ciss{\'{e}}(2018)}]{stock-cisse-2018-convnets}
Pierre Stock and Moustapha Ciss{\'{e}}. 2018.
\newblock \href {https://doi.org/10.1007/978-3-030-01231-1\_31} {Convnets and
  imagenet beyond accuracy: Understanding mistakes and uncovering biases}.
\newblock In \emph{Computer Vision - {ECCV} 2018 - 15th European Conference,
  Munich, Germany, September 8-14, 2018, Proceedings, Part {VI}}, volume 11210
  of \emph{Lecture Notes in Computer Science}, pages 504--519. Springer.

\bibitem[{Sun et~al.(2021)Sun, Webster, Shah, Wang, and
  Johnson}]{sun-etal-2021-they}
Tony Sun, Kellie Webster, Apu Shah, William~Yang Wang, and Melvin Johnson.
  2021.
\newblock \href {https://arxiv.org/abs/2102.06788} {They, them, theirs:
  Rewriting with gender-neutral english}.
\newblock \emph{arXiv preprint arXiv:2102.06788}.

\bibitem[{Talat et~al.(2021)Talat, Blix, Valvoda, Ganesh, Cotterell, and
  Williams}]{talat-etal-2021-word}
Zeerak Talat, Hagen Blix, Josef Valvoda, Maya~Indira Ganesh, Ryan Cotterell,
  and Adina Williams. 2021.
\newblock \href {https://arxiv.org/abs/2111.04158} {A word on machine ethics: A
  response to jiang et al. (2021)}.
\newblock \emph{arXiv preprint arXiv:2111.04158}.

\bibitem[{Thrush et~al.(2022)Thrush, Tirumala, Gupta, Bartolo, Rodriguez, Kane,
  Gaviria~Rojas, Mattson, Williams, and Kiela}]{thrush-etal-2022-dynatask}
Tristan Thrush, Kushal Tirumala, Anmol Gupta, Max Bartolo, Pedro Rodriguez,
  Tariq Kane, William Gaviria~Rojas, Peter Mattson, Adina Williams, and Douwe
  Kiela. 2022.
\newblock \href {https://aclanthology.org/2022.acl-demo.17} {Dynatask: A
  framework for creating dynamic {AI} benchmark tasks}.
\newblock In \emph{Proceedings of the 60th Annual Meeting of the Association
  for Computational Linguistics: System Demonstrations}, pages 174--181,
  Dublin, Ireland. Association for Computational Linguistics.

\bibitem[{Tzioumis(2018)}]{tzioumis2018demographic}
Konstantinos Tzioumis. 2018.
\newblock \href {https://www.nature.com/articles/sdata201825} {Demographic
  aspects of first names}.
\newblock \emph{Scientific data}, 5(1):1--9.

\bibitem[{Vanmassenhove and Monti(2021)}]{vanmassenhove-monti-2021-gender}
Eva Vanmassenhove and Johanna Monti. 2021.
\newblock \href {https://doi.org/10.18653/v1/2021.gebnlp-1.1} {g{EN}der-{IT}:
  An annotated {E}nglish-{I}talian parallel challenge set for cross-linguistic
  natural gender phenomena}.
\newblock In \emph{Proceedings of the 3rd Workshop on Gender Bias in Natural
  Language Processing}, pages 1--7, Online. Association for Computational
  Linguistics.

\bibitem[{Wang et~al.(2018)Wang, Singh, Michael, Hill, Levy, and
  Bowman}]{wang2018glue}
Alex Wang, Amanpreet Singh, Julian Michael, Felix Hill, Omer Levy, and Samuel
  Bowman. 2018.
\newblock \href {https://aclanthology.org/W18-5446/} {Glue: A multi-task
  benchmark and analysis platform for natural language understanding}.
\newblock In \emph{Proceedings of the 2018 EMNLP Workshop BlackboxNLP:
  Analyzing and Interpreting Neural Networks for NLP}, pages 353--355.

\bibitem[{Wang et~al.(2021)Wang, Liu, Gui, Zhang, Zou, Zhou, Ye, Zhang, Zheng,
  Pang, Wu, Li, Zhang, Ma, Fei, Cai, Zhao, Hu, Yan, Tan, Hu, Bian, Liu, Qin,
  Zhu, Xing, Fu, Zhang, Peng, Zheng, Zhou, Wei, Qiu, and
  Huang}]{wang-etal-2021-textflint}
Xiao Wang, Qin Liu, Tao Gui, Qi~Zhang, Yicheng Zou, Xin Zhou, Jiacheng Ye,
  Yongxin Zhang, Rui Zheng, Zexiong Pang, Qinzhuo Wu, Zhengyan Li, Chong Zhang,
  Ruotian Ma, Zichu Fei, Ruijian Cai, Jun Zhao, Xingwu Hu, Zhiheng Yan, Yiding
  Tan, Yuan Hu, Qiyuan Bian, Zhihua Liu, Shan Qin, Bolin Zhu, Xiaoyu Xing,
  Jinlan Fu, Yue Zhang, Minlong Peng, Xiaoqing Zheng, Yaqian Zhou, Zhongyu Wei,
  Xipeng Qiu, and Xuanjing Huang. 2021.
\newblock \href {https://doi.org/10.18653/v1/2021.acl-demo.41} {{T}ext{F}lint:
  Unified multilingual robustness evaluation toolkit for natural language
  processing}.
\newblock In \emph{Proceedings of the 59th Annual Meeting of the Association
  for Computational Linguistics and the 11th International Joint Conference on
  Natural Language Processing: System Demonstrations}, pages 347--355, Online.
  Association for Computational Linguistics.

\bibitem[{Warstadt et~al.(2019)Warstadt, Singh, and
  Bowman}]{warstadt-etal-2019-neural}
Alex Warstadt, Amanpreet Singh, and Samuel~R. Bowman. 2019.
\newblock \href {https://doi.org/10.1162/tacl_a_00290} {Neural network
  acceptability judgments}.
\newblock \emph{Transactions of the Association for Computational Linguistics},
  7:625--641.

\bibitem[{Waseem et~al.(2017)Waseem, Davidson, Warmsley, and
  Weber}]{waseem-etal-2017-understanding}
Zeerak Waseem, Thomas Davidson, Dana Warmsley, and Ingmar Weber. 2017.
\newblock \href {https://doi.org/10.18653/v1/W17-3012} {Understanding abuse: A
  typology of abusive language detection subtasks}.
\newblock In \emph{Proceedings of the First Workshop on Abusive Language
  Online}, pages 78--84, Vancouver, BC, Canada. Association for Computational
  Linguistics.

\bibitem[{Webster et~al.(2020)Webster, Wang, Tenney, Beutel, Pitler, Pavlick,
  Chen, Chi, and Petrov}]{webster-etal-2020-measuring}
Kellie Webster, Xuezhi Wang, Ian Tenney, Alex Beutel, Emily Pitler, Ellie
  Pavlick, Jilin Chen, Ed~Chi, and Slav Petrov. 2020.
\newblock \href {https://arxiv.org/abs/2010.06032} {Measuring and reducing
  gendered correlations in pre-trained models}.
\newblock \emph{arXiv preprint arXiv:2010.06032}.

\bibitem[{Williams et~al.(2018)Williams, Nangia, and
  Bowman}]{williams-etal-2018-broad}
Adina Williams, Nikita Nangia, and Samuel Bowman. 2018.
\newblock \href {https://doi.org/10.18653/v1/N18-1101} {A broad-coverage
  challenge corpus for sentence understanding through inference}.
\newblock In \emph{Proceedings of the 2018 Conference of the North {A}merican
  Chapter of the Association for Computational Linguistics: Human Language
  Technologies, Volume 1 (Long Papers)}, pages 1112--1122, New Orleans,
  Louisiana. Association for Computational Linguistics.

\bibitem[{Zhao and Gordon(2019)}]{zhao-gordon-2019-inherent}
Han Zhao and Geoffrey~J. Gordon. 2019.
\newblock \href
  {https://proceedings.neurips.cc/paper/2019/hash/b4189d9de0fb2b9cce090bd1a15e3420-Abstract.html}
  {Inherent tradeoffs in learning fair representations}.
\newblock In \emph{Advances in Neural Information Processing Systems 32: Annual
  Conference on Neural Information Processing Systems 2019, NeurIPS 2019,
  December 8-14, 2019, Vancouver, BC, Canada}, pages 15649--15659.

\bibitem[{Zhao et~al.(2019)Zhao, Wang, Yatskar, Cotterell, Ordonez, and
  Chang}]{zhao-etal-2019-gender}
Jieyu Zhao, Tianlu Wang, Mark Yatskar, Ryan Cotterell, Vicente Ordonez, and
  Kai-Wei Chang. 2019.
\newblock \href {https://doi.org/10.18653/v1/N19-1064} {Gender bias in
  contextualized word embeddings}.
\newblock In \emph{Proceedings of the 2019 Conference of the North {A}merican
  Chapter of the Association for Computational Linguistics: Human Language
  Technologies, Volume 1 (Long and Short Papers)}, pages 629--634, Minneapolis,
  Minnesota. Association for Computational Linguistics.

\bibitem[{Zhao et~al.(2018)Zhao, Wang, Yatskar, Ordonez, and
  Chang}]{zhao-etal-2018-gender}
Jieyu Zhao, Tianlu Wang, Mark Yatskar, Vicente Ordonez, and Kai-Wei Chang.
  2018.
\newblock \href {https://doi.org/10.18653/v1/N18-2003} {Gender bias in
  coreference resolution: Evaluation and debiasing methods}.
\newblock In \emph{Proceedings of the 2018 Conference of the North {A}merican
  Chapter of the Association for Computational Linguistics: Human Language
  Technologies, Volume 2 (Short Papers)}, pages 15--20, New Orleans, Louisiana.
  Association for Computational Linguistics.

\bibitem[{Zhao and Bethard(2020)}]{zhao-bethard-2020-berts}
Yiyun Zhao and Steven Bethard. 2020.
\newblock \href {https://doi.org/10.18653/v1/2020.acl-main.429} {How does
  {BERT}{'}s attention change when you fine-tune? an analysis methodology and a
  case study in negation scope}.
\newblock In \emph{Proceedings of the 58th Annual Meeting of the Association
  for Computational Linguistics}, pages 4729--4747, Online. Association for
  Computational Linguistics.

\bibitem[{Zhu et~al.(2015)Zhu, Kiros, Zemel, Salakhutdinov, Urtasun, Torralba,
  and Fidler}]{Zhu_2015_ICCV}
Yukun Zhu, Ryan Kiros, Rich Zemel, Ruslan Salakhutdinov, Raquel Urtasun,
  Antonio Torralba, and Sanja Fidler. 2015.
\newblock \href {https://arxiv.org/abs/1506.06724} {Aligning books and movies:
  Towards story-like visual explanations by watching movies and reading books}.
\newblock In \emph{The IEEE International Conference on Computer Vision
  (ICCV)}.

\bibitem[{Zimman and Hayworth(2020)}]{zimman-etal-2020-we}
Lal Zimman and Will Hayworth. 2020.
\newblock \href
  {https://journals.linguisticsociety.org/proceedings/index.php/PLSA/article/view/4728}
  {How we got here: Short-scale change in identity labels for trans, cis, and
  non-binary people in the 2000s}.
\newblock \emph{Proceedings of the Linguistic Society of America},
  5(1):499--513.

\bibitem[{Zliobaite(2015)}]{zliobaite-2015-relation}
Indre Zliobaite. 2015.
\newblock \href {http://arxiv.org/abs/1505.05723} {On the relation between
  accuracy and fairness in binary classification}.
\newblock \emph{CoRR}, abs/1505.05723.

\bibitem[{Zmigrod et~al.(2019)Zmigrod, Mielke, Wallach, and
  Cotterell}]{zmigrod-etal-2019-counterfactual}
Ran Zmigrod, Sabrina~J. Mielke, Hanna Wallach, and Ryan Cotterell. 2019.
\newblock \href {https://doi.org/10.18653/v1/P19-1161} {Counterfactual data
  augmentation for mitigating gender stereotypes in languages with rich
  morphology}.
\newblock In \emph{Proceedings of the 57th Annual Meeting of the Association
  for Computational Linguistics}, pages 1651--1661, Florence, Italy.
  Association for Computational Linguistics.

\end{thebibliography}

\clearpage

\appendix

\section{Problems with Perturbation Augmentation}\label{sec:turbproblems}

While heuristic approaches have been widely used, they suffer from quality issues, which in turn result in particular demographic attributes being excluded in general. Three axis-attributes are most affected, and we will point to them as exemplars of the general issue: non-binary/underspecified, race/ethnicity-african-american, race/ethnicity-white.

To take an obvious example, English language heuristic demographic perturbation systems have to somehow handle the linguistic fact that gendered pronouns have different forms for each grammatical role in so-called ``standard'' English: both the feminine and the masculine pronouns use the same form for two grammatical functions, but not for the same two: \textit{she, \textbf{her, her}, hers} v. \textit{he, him, \textbf{his, his}}. It is not straightforward for a heuristic system given \textit{her} to determine whether to replace it with \textit{his} or \textit{him}. Put simply, a heuristic system that always maps \textit{her} $\rightarrow$ \textit{him} would fail for an example with a possessive (\textit{unfortunately for \textit{her}, I recently changed \textbf{her} schedule} $\rightarrow$ \textit{unfortunately for \textit{him}, I recently changed \textbf{him} schedule}) and one that maps \textit{her} $\rightarrow$ \textit{his} would fail for an example with an accusative (\textit{unfortunately for \textbf{her}, I recently changed \textbf{her} schedule} $\rightarrow$ \textit{unfortunately for \textbf{his}, I recently changed \textbf{his} schedule}). One might hope that a random selection of mappings could help, but since pronouns are highly frequent in natural language, even that sort of noisy approach would lead to a lot of ungrammatical examples. 

The pronoun situation becomes even more complicated when including non-binary gender, since the most frequent pronoun for non-binary gender affects the verb form as well. For example, if we wanted to replace \textit{he}$\rightarrow\ $\textit{they} in the following example, \textit{the owner came to our table and told us \textbf{he} already \textbf{is} thinking about starting a Turkish breakfast}, this would result in another grammatically incorrect sentence, \textit{the owner came to our table and told us \textbf{they} already \textbf{is} thinking about starting a Turkish breakfast}. One might hope that one could just add bigrams to the word lists containing pronouns and all verb forms, but that doesn't straightforwardly work, as other words (sometimes several of them) can appear between the pronoun and the verb, and thus not be caught by a heuristic system. Although this particular issue only occurs (in English) in the context of singular \textit{they}, it would be counter to the goals of a responsible AI work such as this one to accept higher noise for underserved identities like non-binary that are often ignored or overlooked in NLP tasks \citep{sun-etal-2021-they, lauscher-etal-2022-welcome}.

As if the situation with pronouns weren't complicated enough, often context is needed to determine whether particular words should be perturbed at all. For example, ``Black'' and ``white'' are polysemous adjectives that can be used not only as demographic terms but also as color terms. Despite the fact that these references aren't demographic, they would get perturbed by nearly every heuristic demographic perturbation system (\textit{the person was wearing \textbf{a white} shirt} $\rightarrow$ \textit{the person was wearing \textit{an Asian} shirt} or \textit{the \textbf{white} pawn attacked the black bishop} $\rightarrow$ \textit{the black pawn attacked the black bishop}), altering the meaning significantly. If a heuristic system like this were used to measure model robustness to demographic perturbation say in an NLU classification task like natural language inference, it would be hard to determine whether the model failed to be robust to demographic changes (and hence should be deemed unfair) or if the textual changes had altered the meaning too much and that affected the label. 

% \section{Demographic Axes and Attributes}
% In this section, we describe the demographic axes and attributes used in \acronym~data collection and perturbation augmentation. We follow the 2020 US Census Survey for race/ethnicity categories. For age, we used bucketed age groups, due to the different formats age information is presented (\textit{twenty years old} vs. \textit{young man}). We provide an additional category of Adult (unspecified), given the frequency of under-specified age descriptors in language, e.g., \textit{man, father}.

% The full list of demographic axes and attributes are presented below:
% \begin{itemize}[noitemsep,nolistsep]
%     \item Gender
%         \begin{itemize}
%             \item Man
%             \item Woman
%             \item Non-Binary/Underspecified
%         \end{itemize}
%     \item Race/Ethnicity (U.S. Census)
%     \begin{itemize}
%         \item White
%         \item Black
%         \item Hispanic or Latino
%         \item Asian
%         \item Native American or Alaska Native
%         \item Hawaiian or Pacific Islander
%     \end{itemize}
%     \item Age
%     \begin{itemize}
%         \item Child (< 18)
%         \item Young (18-44)
%         \item Middle-aged (45-64)
%         \item Senior (65+)
%         \item Adult (Unspecified)
%     \end{itemize}
%     \label{tab:axis-attributes}
% \end{itemize}
 
\section{Data Collection Task Layout}\label{sec:annotationdetails}

% \begin{figure*}[ht] %placeholder fig. sketch
%     \includegraphics[width=0.95\linewidth]{crowdsourcing.png}
%     \caption{Three Stages of Crowdsourced Data Collection}
%     \label{fig:datacollection}
% \end{figure*}

\begin{figure*}[h!] %placeholder fig. sketch
      \begin{center}
\begin{adjustbox}{width=\linewidth}
    \includegraphics[width=0.8\textwidth]{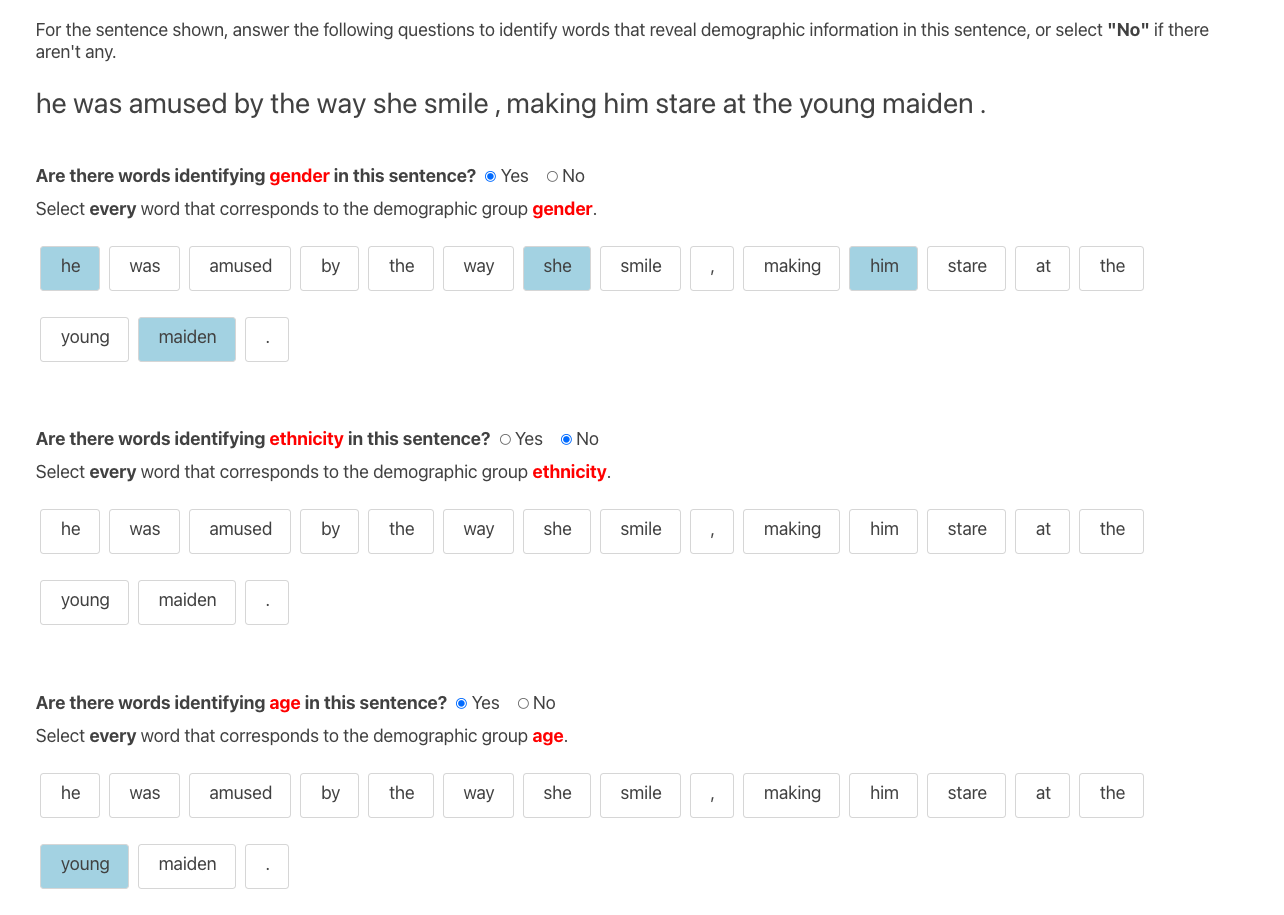}
    \end{adjustbox}
    \end{center}
    \caption{Design of Stage 1 of data collection, in which annotators select demographic terms in a text snippet.}
    \label{fig:datacollectionphase1}
\end{figure*}

We collected \acronym~over three stages, each with a different crowdworker participating:

\paragraph{Stage 1:} A crowdworker is presented with a snippet that our preprocessing stage indicated was probably perturbable. They select perturbable words in the snippet by demographic axis (gender, age, race). Crowdworkers often select words that were used during preprocessing for perturbability scoring (for the scoring function see \autoref{eq:perturbability-score}). Employing humans in this stage also enabled us to filter out examples that were erroneously flagged as perturbable during preprocessing. % (recall \S\ref{subsec:problemsheuristic}).
See the annotation interface for Stage 1 in \autoref{fig:datacollectionphase1}.

\begin{figure*}[h!] %placeholder fig. sketch
      \begin{center}
\begin{adjustbox}{width=\linewidth}
    \includegraphics[width=0.8\textwidth]{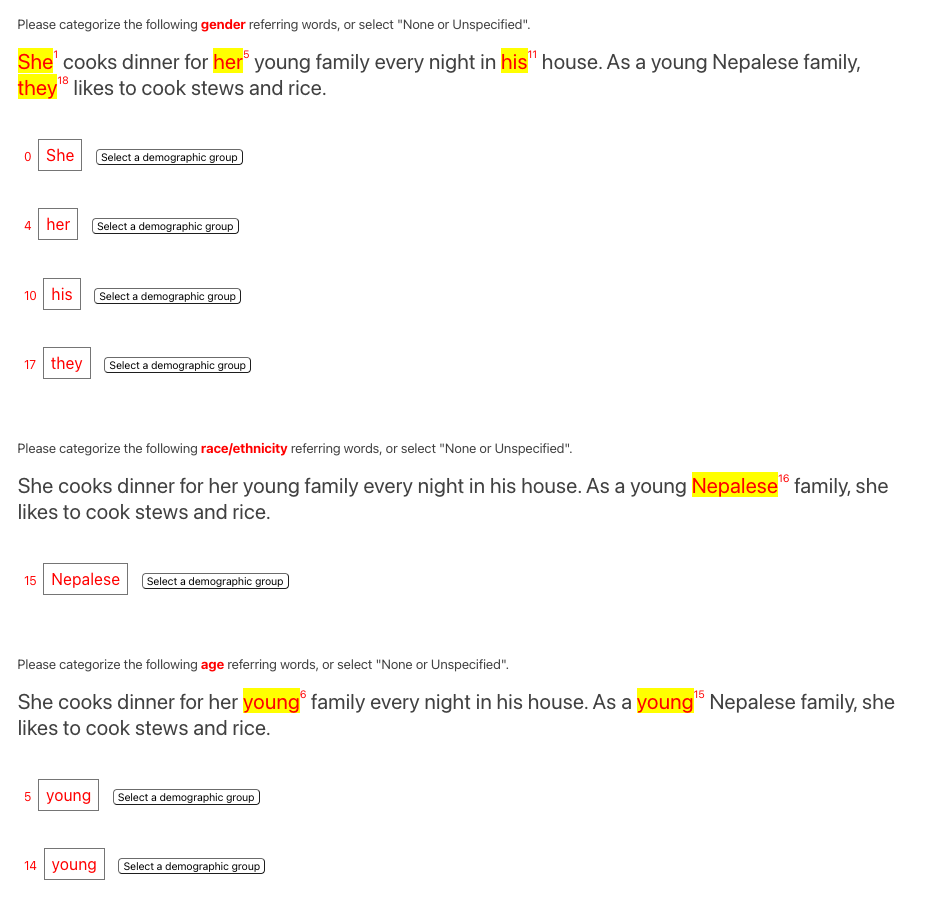}
    \end{adjustbox}
    \end{center}
    \caption{Design of Stage 2 of data collection, in which annotators assign attributes to demographic words selected during Stage 1.}
    \label{fig:datacollectionphase2}
\end{figure*}

\paragraph{Stage 2:} A crowdworker is presented with a text snippet that Stage 1 determined to contain one or more words associated with a demographic axis (gender, race, and age). For each selected word, the worker chooses from a drop-down menu which particular demographic attribute the word instantiates: for example, all words highlighted in Stage 1 as being perturbable along the gender axis are labeled either as referring to a ``man'', ``woman'', or someone who is ``non-binary/underspecified''. This enables better treatment or coreference resolution, as humans will be able to determine which perturbable words refer to the same person better than a heuristic system could. See the annotation interface for Stage 2 in \autoref{fig:datacollectionphase2}.

\begin{figure*}[h!] %placeholder fig. sketch
        \begin{center}
\begin{adjustbox}{width=\linewidth}
    \includegraphics[width=0.8\textwidth]{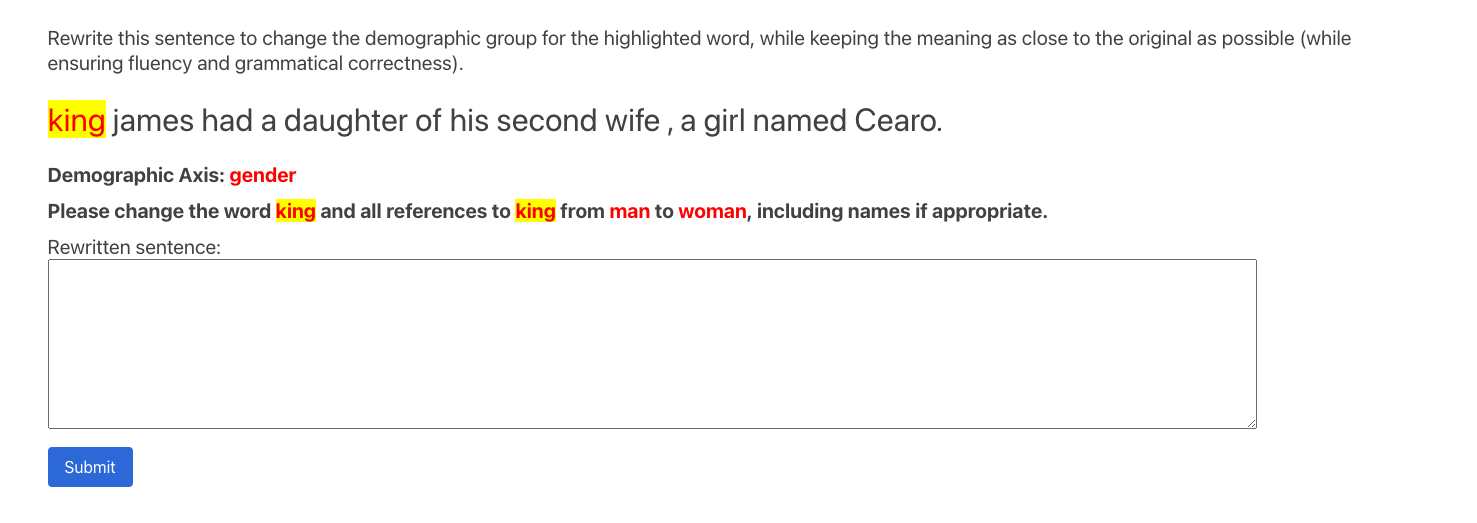}
    \end{adjustbox}
    \end{center}
    \caption{Design of Stage 3 of data collection, in which annotators rewrite the text by changing the demographic attribute of all references to the selected word, while preserving meaning and fluency.}
    \label{fig:datacollectionphase3}
\end{figure*}

\paragraph{Stage 3:} Given a text snippet $s$, highlighted perturbable word $w \in s$ and $source$ and $target$ attributes, a crowdworker creates a minimal distance re-write of text snippet $\widetilde{s}$. %See \autoref{fig:datacollection} for a schematic of our three step crowdsourcing process for collecting human perturbed data for training the \perturber.
See the annotation interface for Stage 3 in \autoref{fig:datacollectionphase3}.

% annotator demographics:
\section{Annotator Demographics}\label{app_sec:annotators}

Recent works have questioned the ethics of releasing crowdsourced datasets without reporting demographic information for human annotators who create the annotations (\citealt{bender-friedman-2018-data}). Given the complexity and potential subjectivity of the demographic perturbation task, we believe it is especially important to examine the demographic make-up of our annotator pool, and maintain open channels of communication with our crowd-workers. In our annotation tasks, we included an opt-in demographic survey after task completion that allowed workers to report their gender, ethnicity and age. Whether a worker chose to participate in the demographic survey did not affect their payment.

Prior demographic surveys of MTurk workers often exclude historically marginalized groups \citep{moss-etal-2020-demographic}, such as non-binary people and Native-American people, who we include. Of our survey responses, 0.7\% identified as non-binary, and 2.1\% identified as Native American, suggesting that prior analysis of worker pools do not reveal the full spectrum of identities. 

For gender identity, our annotations were performed primarily by people who self-identified as Woman (28.4\%), Man (24.7\%), Woman/Non-Binary (0.6\%), or Non-Binary (0.1\%). Additional gender identities consisted of $\leq$ 100 annotations each, and 46.1\% of annotations were performed by people who opted out of the gender portion of the survey. For race, our annotations were performed primarily by people self-identifying as White (38.4\%), Hispanic or Latinx (7.0\%), Black (2.3\%), Native American (2.1\%), Asian (2.1\%), Hispanic and White (0.7\%), or Asian and White (0.2\%). Additional racial identities consisted of $\leq$ 10 workers each, and 4.4\% of the workers declined to respond. For the annotations that received responses to the age portion of the survey (52748/98583), the mean age was 38.6 years and the median was 36 years, with a standard deviation of 10.2 years. 

\begin{table*}[h!]
    \centering
    \footnotesize
    \begin{tabular}{lrrrr}
        \toprule
        \bf race & \bf \# annotations & \bf \% annotations & \bf \# annotations$^{\alpha}$ & \bf \% annotations$^{\alpha}$ \\
        \midrule
        no answer                                               &          46350 &           47.0 &                      42420 &                       61.9 \\
        white                                                   &          37887 &           38.4 &                      21443 &                       31.3 \\
        hispanic                                                &           6907 &            7.0 &                       1274 &                        1.9 \\
        black                                                   &           2253 &            2.3 &                       1142 &                        1.7 \\
        native-american                                         &           2089 &            2.1 &                          0 &                        0.0 \\
        asian                                                   &           2058 &            2.1 &                       1670 &                        2.4 \\
        hispanic, white                                         &            730 &            0.7 &                        454 &                        0.7 \\
        asian, white                                            &            212 &            0.2 &                   $\leq$ 100 &                          X \\
        black, hispanic                                         &       $\leq$ 100 &              X &                   $\leq$ 100 &                          X \\
        native-american, white                                  &       $\leq$ 100 &              X &                   $\leq$ 100 &                          X \\
        pacific                                                 &       $\leq$ 100 &              X &                          0 &                        0.0 \\
        asian, black, hispanic, native-american, pacific, white &       $\leq$ 100 &              X &                          0 &                        0.0 \\
        black, white                                            &       $\leq$ 100 &              X &                          0 &                        0.0 \\
        asian, hispanic                                         &       $\leq$ 100 &              X &                          0 &                        0.0 \\
        \midrule
        \bf total                                                   &          98583 &          100.0 &                      68524 &                      100.0 \\
        \bottomrule
        %\rule{0pt}{0.5\normalbaselineskip}
        %& \rule{0pt}{0.1\normalbaselineskip}
        %& \rule{0pt}{0.1\normalbaselineskip}
        %& \rule{0pt}{0.1\normalbaselineskip}
        %& \rule{0pt}{0.1\normalbaselineskip}\\
        
        \toprule
        \bf gender & \bf \# annotations & \bf \% annotations & \bf \# annotations$^{\alpha}$ & \bf \% annotations$^{\alpha}$ \\
        \midrule
        no answer                       &          45475 &           46.1 &                      41591 &                       60.7 \\
        woman                           &          27965 &           28.4 &                      16155 &                       23.6 \\
        man                             &          24329 &           24.7 &                      10128 &                       14.8 \\
        woman, non-binary               &            614 &            0.6 &                        613 &                        0.9 \\
        non-binary                      &            146 &            0.1 &                   $\leq$ 100 &                          X \\
        man, non-binary                 &       $\leq$ 100 &              X &                   $\leq$ 100 &                          X \\
        woman, man                      &       $\leq$ 100 &              X &                   $\leq$ 100 &                          X \\
        woman, man, non-binary, other   &       $\leq$ 100 &              X &                          0 &                        0.0 \\
        other                           &       $\leq$ 100 &              X &                          0 &                        0.0 \\
        woman, man, non-binary          &       $\leq$ 100 &              X &                          0 &                        0.0 \\
        \midrule
        \bf total                           &          98583 &          100.0 &                      68524 &                      100.0 \\
        \bottomrule
        
    \end{tabular}
     \caption{All responses to the race (top) and gender (bottom) surveys.  Responses with 100 or fewer instances have been obscured to protect worker identities. Columns marked with $^{\alpha}$ denote allowlist annotations that were were done by workers who demonstrated high quality work and were tasked with annotating the majority of the dataset (68524 out of 98583 examples). }
    \label{tab:turkerrace-gender}
\end{table*}

\begin{table}[h]
    \begin{center}
\centering
    \small
    \begin{tabular}{lrr}
    \toprule
    {} &   annotations &   annotations$^{\alpha}$  \\
    \midrule
    mean      &         38.6 &                     41.5 \\
    median    &         36.0 &                     42.0 \\
    std       &         10.2 &                     10.3 \\
    min       &           18 &                       19 \\
    max       &           80 &                       80 \\
    \midrule
    responses &        52748 &                    26867 \\
    size      &        98583 &                    68524 \\
    \bottomrule
    \end{tabular}
    \end{center}
        \caption{Age statistics across all annotations. The last two rows show the number of responses received and total annotations. Columns marked with $^{\alpha}$ refer allowlist annotations which were were done by workers who demonstrated high quality work and were tasked with annotating the majority of the dataset (68524 out of 98583 examples). }
    \label{tab:turkerage}
\end{table}

\section{Inter-Annotator Agreement}\label{sec:iaa}
As described in the main text, we calculated inter-annotator agreement metrics across rewrites of NLI premises through naive token and entire annotation level agreement, Levenshtein distance, and various other traditional metrics (see \autoref{tab:iaa-summary}). Annotation level agreement was calculated by isolating exact matches between rewrites, and returning the proportion of them that belong to the majority. Token level agreement was calculated by isolating exact matches at each token position, returning the proportion of tokens at that position which belong to the majority, and then taking the mean score across the entire annotation. For all our other metrics (sacreBLEU, ROUGE1, ROUGE2, ROUGE-L, ROUGE-L$_{\text{sum}}$, Levenshtein Distance), we calculated pairwise scores in both directions, and then took the mean across all scores.

\begin{table*}
    \begin{center}
\begin{adjustbox}{width=\linewidth}
    \begin{tabular}{rrrrrrrr}
    \toprule
    \bf full \% agree &  \bf token \% agree &  \bf sacreBLEU &  \bf ROUGE1 &  \bf ROUGE2 &  \bf ROUGE-L &  \bf ROUGE-L\textsubscript{sum} &  \bf Levenshtein \\
    \midrule
    73.65 &                  94.84 &      92.85 &   94.44 &   89.92 &   94.32 &      94.32 &         0.06 \\
    \bottomrule
    \end{tabular}
\end{adjustbox}
\end{center}
    \caption{Summary of inter-annotator agreement metrics. Tasks where only one annotation was available were excluded as they have trivially perfect agreement scores, and the mean score across all remaining tasks is reported here.}
    \label{tab:iaa-summary}
\end{table*}

\section{Data Quality Hand Audit}\label{sec:dataaudit}
We randomly selected 300 examples to be annotated by 4 expert annotators; each example was annotated by 2 experts in order to get an estimate of interannotator agreement. To get an idea of the types of errors that affect our dataset, we recruited four experts to contribute to a dataset audit. Our annotation scheme followed work by \citet{blodgett-etal-2021-stereotyping} which urges dataset audits and highlighted pitfalls of fairness dataset creation. We will release their anonymized annotations along with the other artifacts from this work. The experts contributed a hand annotation of 300 example snippets from \acronym, the results of which are reported in \autoref{tab:data-quality-summary}. 

The imperfections uncovered through the dataset audit vary in their severity. Some represent common annotation issues that are prevalent in crowdsourced human data generation (e.g., typos), others are a direct consequence of our counterfactual methodological approach (e.g., factuality changes), while still others (e.g.,  perturbations to the wrong attribute, and sensitive factual changes) provide an initial estimate of the noisiness of \acronym. 

We have chosen to be liberal in reporting imperfections in \autoref{tab:data-quality-summary}, since we believe that any dataset noise could be potentially problematic, and we would like to be as transparent as possible about any data issues. Many of the examples that were found contain imperfections are still useable and do not contain offensive content. This being said, the work outlined in this section represents a preliminary dataset audit on a very small portion of \acronym---we plan to continue to explore \acronym, describe its contents more thoroughly, and quantify noise and other dataset issues going forward. We encourage other researchers to share any issues if they find them, and to be circumspect in how they use the research artifacts we describe here. 

Annotators were tasked with identifying the following imperfections:

\begin{table}[h!]
    \centering
    \small
    \begin{tabular}{lrr}
    \toprule
    \it Tag & \it  \% occurrence & \it  \% agreement \\
    \midrule
    factuality change                        &          44.7 &         74.0 \\
    % sensitive factuality change              &           5.7 &         94.3 \\ 
    % \midrule
    incomplete/incorrect                  &          25.7 &         84.0 \\
    Stage 1 errors &          18.7 &         88.0 \\
    typos and naturalness                            &          18.0 &         87.3 \\
    incorrectly unperturbed                 &           3.7 &         96.7 \\
    \bottomrule
    \end{tabular}
        \caption{Expert dataset audit. Under each tag, we report the rate at which it occurs, as well as how often two annotators agreed. If either annotator included a tag for an example, that tag was aggregated as \% occurrence.}
    \label{tab:data-quality-summary}
\end{table}

\paragraph{Typos and Naturalness:} This category was investigated to uncover general data quality issues. Most of these issues resulted from annotators misunderstanding the task, or from the source text. We investigated four possible issues related to perturbation annotations affecting the grammar and the textual flow in snippets, and provide examples of issues we found below:
    \begin{itemize}
        \item \textbf{Grammatical error}: \textit{What book did Frederick Lynch author? Frederick Lynch, the author of Invisible \textbf{Victims}: White Males and the Crisis of Affirmative Action, did a study on white males that said they were \textbf{victims} of reverse discrimination\textellipsis} $\rightarrow$ \textit{What book did Frederick Lynch author ? Frederick Lynch , the author of Invisible \textcolor{red}{\textbf{Young (18-44)}} : White Males and the Crisis of Affirmative Action , did a study on white males that said they were \textcolor{red}{\textbf{Young (18-44)}} of reverse discrimination\textellipsis}
        \item \textbf{New typos introduced}: \textit{simon now worried that he would not be able to make it to the plane in time , slowly walked towards the voice , hoping his martial arts training and the adrenaline he felt would be enough for what \textbf{he was} going to do next .} $\rightarrow$	\textit{Jordan now worried that they would not be able to make it to the plane in time , slowly walked towards the voice , hoping their martial arts training and the adrenaline they felt would be enough for what \textcolor{red}{\textbf{there were}} going to do next .}
        \item \textbf{Unnecessary word insertion} \textit{for her tenth birthday, after she had ceremonially burned her dolls and all things girly a week before, she finally beat her parents (father) into submission and got her first dinghy.} $\rightarrow$ \textit{for her tenth birthday, after she had ceremonially burned her dolls and all things girly a week before, she finally beat her parents (\textcolor{red}{\textbf{non-birthing}} parent) into submission and got her first dinghy.}
        \item \textbf{Marking of a group that wouldn't normally be marked \cite{blodgett-etal-2021-stereotyping}}: ...\textit{Her parents were executed via guillotine by the Zanscare Empire.} $\rightarrow$ ...\textit{Her \textcolor{red}{\textbf{young adult}} parents were executed via guillotine by the Zanscare Empire.}
    \end{itemize}

\paragraph{Incomplete or Incorrect perturbation:} These arise where the perturbation wasn't correctly applied. The most common type of incorrect perturbation was failing to perturb the entire coreference chain, although such examples still yield partial signal for the \perturber~to learn from. 
    \begin{itemize}
        \item \textbf{Failure to perturb the entire co-reference chain}: It is cognitively taxing to trace and alter every pronoun in a long reference chain, and sometimes annotators failed to catch every perturbable pronoun. Often these are  examples where it is ambiguous whether a pronoun appears on the chain or not, such as \textit{he saw his cat}---do the two masculine pronouns refer to one person or two? It is hard to tell in a sentence with little provided context, and there can even be some variation for longer sentences like the following example: \textit{\textellipsis In \textbf{his} second year \textbf{he} neglected his medical studies for natural history and spent four months assisting Robert Grant's research into marine invertebrates. \textellipsis Filled with zeal for science, \textbf{he} studied catastrophist geology with Adam Sedgwick.} $\rightarrow$	\textit{\textellipsis In \textcolor{green}{her} second year \textcolor{red}{he} neglected \textcolor{red}{his} medical studies for natural history and spent four months assisting Robert Grant 's research into marine invertebrates. \textellipsis  Filled with zeal for science , \textcolor{red}{he} studied catastrophist geology with Adam Sedgwick .}
        \item \textbf{Perturbation of entities not on the co-reference chain:} We worried that people would perturb relations that weren't on the coreference chain, despite being instructed against it, for examples such as \textit{she saw her husband} $\rightarrow$ \textit{he saw his \textcolor{red}{wife}}, according to preconceptions about stereotypes like heteronormativity. The expert annotators  didn't observe snippets with these errors. In the expert annotated sample, the majority of these examples were ones where the annotator of Stage 3 fixed typos (such as lack of capitalization on the first word) that cascaded through data collection from the source.
        \item \textbf{Perturbation to wrong demographic group}: Sometimes workers would perturb an example to a demographic group not specified by the task. In this example the worker was instructed to perturb from woman to man, but instead perturbed from woman to non-binary: \textit{To herself she said: "Of course, if father heard that he would have a fit! She thought to herself: ``Father would be fine with that.''} $\rightarrow$ \textit{To \textcolor{red}{\textbf{themself}} \textcolor{red}{\textbf{they}} said: ``Of course, if father heard that he would have a fit! \textcolor{red}{\textbf{They}} thought to \textcolor{red}{\textbf{themself}}: ``Father would be fine with that.''}
        \item \textbf{Perturbation of words unnecessarily}: Perturbation of words that don't convey demographic information, such as surname when the demographic axis is gender. \textit{Affirms the gifts of all involved, starting with Spielberg and going right through the ranks of the players -- on-camera and off -- that he brings together.} $\rightarrow$ \textit{Affirms the gifts of all involved, starting with \textcolor{red}{\textbf{Jenkins}} and going right through the ranks of the players -- on-camera and off -- that she brings together.}
        \item \textbf{Perturbing names to pronouns}: Occasionally, names would be replaced with pronouns---the result was usually grammatical and the gender axis was perturbed as instructed, but the perturbation isn't perfect, for example \textit{Peralta's mythmaking could have used some informed, adult hindsight.} $\rightarrow$ \textit{\textcolor{red}{\textbf{Their}} mythmaking could have used some informed, adult hindsight.}
    \end{itemize}
\paragraph{Errors from Stage 1:} Often there was an annotation error from data collection Stage 1 (Word Identification) that lead to an unusual word being presented as perturbable in later stages of data collection when it shouldn't have been. These issues generally don't result in errors down the line, since Stage 3 annotators catch these errors, but a few made it into \acronym. The expert annotators looked for two types of errors from Stage 1:  %{Incorrect word identification leading to problematic perturbation}
    \begin{itemize}
        \item \textbf{Chosen word doesn't refer to a person}: In one example an annotator highlighted the word \textit{questions} as a gender-perturbable word in \textit{\textellipsis it is possible to answer these \textcolor{red}{\textbf{questions}} purely within the realm of science\textellipsis}.
        \item \textbf{Chosen word doesn't refer to the intended demographic axis}: For example, in the snippet \textit{While certainly more naturalistic than its \textcolor{red}{\textbf{Australian}} counterpart, Amari's film falls short in building the drama of Lilia's journey}, \textit{Australian} was selected as perturbable along race, but this word actually refers to a nationality with citizens of numerous races/ethnicities. For example, perturbing \textit{Australian} to \textit{Black} presupposes that the two sets are disjoint, when they may actually overlap and represents an error. These errors were relatively common, since there are strong and often stereotypical statistical associations between nationalities and race/ethnicity that can be hard for untrained crowdsourced annotators to avoid. See \newcite{blodgett-etal-2021-stereotyping} for a related observation. 
    \end{itemize}
\paragraph{Incorrectly unperturbed:} Some perturbable examples had correct word identification, but were left unperturbed. For this example, workers were asked to perturb a child to an adult, but the age information in the example was left unperturbed: \textit{BC. Our two year old granddaughter came to Boston last weekend. Her mother and father went to visit Boston College. They went to school there in 2003-2007. They bought her a BC t-shirt. She looked cute in it. They went to school there in 2003-2009.}

\paragraph{Factuality changes:} These errors occur when perturbing a demographic reference changes a known fact about the real world to an alternative counterfactual version. 

    \begin{itemize}
        \item \textbf{Perturbation changes facts about known entities}: see the examples in \autoref{tab:datasetexamples} about \textit{King Victor} and \textit{Asian Austin Powers}.
        \item \textbf{Perturbation invents new terms/phrases}: \textit{Brady achieves the remarkable feat of squandering a topnotch foursome of actors... by shoving them into every clich\'ed white-trash situation imaginable.} $\rightarrow$ \textit{Brady achieves the remarkable feat of squandering a topnotch foursome of actors... by shoving them into every clich\'ed \textcolor{red}{\textbf{black-trash}} situation imaginable.}
    \end{itemize}

\paragraph{Removal of offensive examples:} The factuality changes described above are often benign, but in the annotation process experts became concerned that we had a few examples where the nature of the factuality issue could cause harm or offense to particular demographic groups. To explore these we applied an automatic toxicity classifier \citep{dinan-etal-2019-build} to try to find offensive examples. However, we found that the classifier predominantly flagged examples that contained explicit themes, but often that these examples weren't necessarily harmful to a particular group according to our experts. For example, the rewrite \textit{to his great surprise, he removed his hood to reveal a bloody face, scarred beyond human recognition} $\rightarrow$ \textit{to their great surprise, they removed their hood to reveal a bloody face, scarred beyond human recognition} was deemed ``offensive'' by the classifier, but is actually a good example that we want to keep in our dataset.

In the absence of a clear automatic way to detect harmful perturbations, we opted to apply dataset filtering judiciously. We made the judgment call to remove examples that caused the most direct harm to racial minorities through a manual review of the changed noun phrases between unperturbed and perturbed texts. In this process, we targeted two major sources of harm caused by perturbation: (i) the creation of new slurs targeting current racial minorities in the United States e.g., \textit{white-trash} $\rightarrow$ \textit{black-trash}, and (ii) the positioning of historically oppressed groups as oppressors e.g., \textit{white supremacy} $\rightarrow$ \textit{cherokee supremacy}. We then selected 50 perturbed phrases %(see Table \ref{tab:offensive-phrases}) 
 containing one of these harms, and removed a total of 43 examples containing these phrases from the dataset. There are likely other instances of offensive content in \dataset \ that are yet to be discovered, but we hope to have removed at least some of the most egregious.

\section{Perturber Training Parameters}

In this section, we describe hyperparameters for training the \perturber.
\autoref{tab:perturber-training-params} describes the hyperparameters for finetuning BART-Large \citep{lewis-etal-2020-bart} on \acronym, with 24 layers, 1024 hidden size, 16 attention heads and 406M parameters. Validation patience refers to the number of epochs where validation loss does not improve, used for early stopping. All perturber training and evaluation runs are conducted using the ParlAI library \citep{miller-etal-2017-parlai}.\footnote{\url{https://parl.ai}} We trained the \perturber~using 8 $\times$ 16GB Nvidia V100 GPUs for approximately 4 hours.

\begin{table}[h]
    \begin{center}
%\begin{adjustbox}{width=\linewidth}
\addtolength{\tabcolsep}{0pt}
\renewcommand{\arraystretch}{1.0}
    %\centering\small\renewcommand{\arraystretch}{1}
    \begin{footnotesize}
    \begin{tabular}{lr}
    \toprule
    \bf Hyperparam & \bf PANDA \\\midrule
    Learning Rate     & 1e-5 \\
    Batch Size   & 64 \\
    Weight Decay     & 0.01 \\
    Validation Patience & 10 \\
    Learning Rate Decay & 0.01 \\
    %Warmup Ratio & 0.0001 \\
    Warmup Updates   & 1200 \\
    Adam $\epsilon$  & 1e-8 \\
    Adam $\beta_1$  & 0.9 \\
    Adam $\beta_2$ & 0.999 \\
    Gradient Clipping   & 0.1 \\
    Decoding Strategy & greedy \\
    \bottomrule
    \end{tabular}
    \end{footnotesize}
    %\end{adjustbox}
    \end{center}
    \caption{Hyperparameters for training the perturber by finetuning BART on \acronym.}
    \label{tab:perturber-training-params}
\end{table}

\section{FairBERTa Training Parameters}

\autoref{tab:fairberta-training-params} contains hyperparameters for pretraining \fairberta. \fairberta~is trained with the RoBERTa$_{\textrm{BASE}}$ \cite{liu-etal-2019-roberta} architecture on 32GB Nvidia V100 GPUs with mixed precision using the Fairseq library \cite{ott2019fairseq}. We pretrain \fairberta~on 160GB perturbed data using 256 V100 GPUs for approximately three days. For RoBERTa and \fairberta~models trained on the 16GB BookWiki corpus (and perturbed BookWiki corpus), we use the same training settings, but use 100K max steps.
%equaling 125M total parameters.

\begin{table}[h]
    \begin{center}
%\begin{adjustbox}{width=\linewidth}
\addtolength{\tabcolsep}{0pt}
\renewcommand{\arraystretch}{1.0}
\footnotesize
    %\centering\small\renewcommand{\arraystretch}{1}
    \begin{footnotesize}
    \begin{tabular}{lr}
    \toprule
    \bf Hyperparam & \bf \fairberta \\\midrule
    \# Layers     & 12 \\
    Hidden Size     & 768 \\
    FFN inner Hidden Size     & 3072 \\
    \# Attention Heads     & 12 \\
    Attention Head Size     & 64 \\
    Hidden Dropout     & 0.1 \\
    Attention Dropout   & 0.1 \\
    \# Warmup Steps     & 24k \\
    Peak Learning Rate     & 6e-4 \\
    Batch Size & 8k \\
    Weight Decay    & 0.01 \\
    Sequence Length & 512 \\
    Max Steps & 500k \\
    Learning Rate Decay & Linear \\
    Adam $\epsilon$  & 1e-8 \\
    Adam $\beta_1$  & 0.9 \\
    Adam $\beta_2$ & 0.999 \\
    Gradient Clipping & 0.0 \\
    \bottomrule
    \end{tabular}
    \end{footnotesize}
   % \end{adjustbox}
    \end{center}
    \caption{Hyperparameters for pretraining \fairberta.}
    \label{tab:fairberta-training-params}
\end{table}

\section{Downstream Task Training Parameters}

\autoref{tab:glue-training-params} describes hyperparameters for finetuning and fairtuning RoBERTa and \fairberta~on GLUE tasks and the RACE \citep{lai-etal-2017-race} reading comprehension dataset. We conducted a basic hyperparameter exploration sweeping over learning rate and batch size, and select the best hyperparameter values based on the median validation accuracy of 3 runs for each task. Configurations for individual models, tuning approach and GLUE task will be released in our GitHub repository. Training runs on downstream tasks are done using HuggingFace. Models are trained on 8 $\times$ 32GB Nvidia V100 machines, with runtime ranging from 5 minutes for the smallest dataset (RTE) to 45 minutes for the largest dataset (QQP).

\begin{table}[h]
    \begin{center}
%\begin{adjustbox}{width=\linewidth}
\addtolength{\tabcolsep}{0pt}
\renewcommand{\arraystretch}{1.0}
\begin{footnotesize}
    %\centering\small\renewcommand{\arraystretch}{1}
    \begin{tabular}{lrr}
    \toprule
    \bf Hyperparam & \bf GLUE & \bf RACE \\\midrule
    Learning Rate     & \{1e-5, 2e-5, 3e-5\} & 1e-5 \\
    Batch Size   & \{16, 32\} & 16 \\
    Weight Decay     & 0.1 & 0.1 \\
    Max \# Epochs & 10 & 3 \\
    Learning Rate Decay & Linear & Linear \\
    Warmup Ratio & 0.06 & 0.06 \\
    \bottomrule
    \end{tabular}
    \end{footnotesize}
 %   \end{adjustbox}
    \end{center}
    \caption{Hyperparameters for finetuning RoBERTa and \fairberta~on GLUE and RACE.}
    \label{tab:glue-training-params}
\end{table}

\section{Additional GLUE Statistics}\label{sec:downstreamperformance}

We provide the percentage of examples in the validation set (used for reporting accuracy as test sets are hidden) that were perturbed across six tasks from the GLUE benchmark in \autoref{tab:glue-stats}. CoLA and RTE had the highest percentage of perturbable examples, followed by QNLI and STS-B, with SST-2 having the fewest. 

\begin{table}[h]
    \begin{center}
\begin{adjustbox}{width=\linewidth}
\addtolength{\tabcolsep}{0pt}
\renewcommand{\arraystretch}{1.0}
\footnotesize
    %\centering\small\renewcommand{\arraystretch}{1}
    \begin{tabular}{lrrrrrr}
    \toprule
     & \bf CoLA & \bf SST-2 & \bf STS-B & \bf QQP & \bf RTE & \bf QNLI \\\midrule
    \it age      & 9.2 & 7.5  & 12.2  & 6.4 & 13.2 & 6.5 \\
    \it gender   & 32.3 & 9.9  & 20.2 & 8.3 & 32.9 & 18.4 \\
    \it race     & 4.1  & 5.2  & 4.5  & 5.8 & 0.8 &  6.3 \\\midrule
    \it total & 45.6 & 22.5 & 36.9 & 20.5 & 47 & 31.2 \\
    \bottomrule
    \end{tabular}
    \end{adjustbox}
    \end{center}
    \caption{The percentage of examples perturbed by demographic axis for each fairtuning task.}
    \label{tab:glue-stats}
\end{table}

\begin{table*}[h!]
  \begin{center}
  \footnotesize
  %\begin{adjustbox}{width=\linewidth}
  \begin{tabular}{lp{10cm}ll}
    \toprule
        Dataset & Input & Label & Perturbed \\
        \midrule
        RTE & \textbf{premise}: \textit{Swansea striker Lee Trundle has negotiated a lucrative image-rights deal with the League One club.} \textbf{hypothesis}: \textit{Lee Trundle is in business with the League One club.} & entailment & No \\
        RTE	& \textbf{premise}: \textit{Swansea striker \textcolor{blue}{Lisa} Trundle has negotiated a lucrative image-rights deal with the League One club.} \textbf{hypothesis}: \textit{\textcolor{blue}{Lisa} Trundle is in business with the League One club.} & entailment & Yes \\
        \midrule
        SST-2 & \textit{his healthy sense of satire is light and fun ...} & positive & No \\
        SST-2 & \textit{\textcolor{blue}{their} healthy sense of satire is light and fun ...} & positive	& Yes \\
        \midrule
        QNLI & \textbf{question}: \textit{How many people lived in Warsaw in 1939?} \textbf{sentence}: \textit{Unfortunately this belief still lives on in Poland (although not as much as it used to be)} & not entailment & No \\
        QNLI & \textbf{question}: \textit{How many \textcolor{blue}{women} lived in Warsaw in 1939?} \textbf{sentence}: \textit{Unfortunately this belief still lives on in Poland (although not as much as it used to be)} & not entailment & Yes \\
        \midrule
        QQP & \textbf{question 1}: \textit{Do women cheat more than men?} \textbf{question 2}: \textit{Do more women cheat than men?} & not duplicate & No \\
        QQP & \textbf{question 1}: \textit{Do \textcolor{blue}{middle-aged} women cheat more than men?} \textbf{question 2}: \textit{Do more \textcolor{blue}{middle-aged} women cheat than men?} & not duplicate & Yes \\
        \midrule
        CoLA & \textit{John arranged for himself to get the prize.} & acceptable & No \\
        CoLA & \textit{\textcolor{blue}{Joanne} arranged for \textcolor{blue}{herself} to get the prize.} & acceptable & Yes \\
        \midrule
        STSB & \textbf{sentence 1}: \textit{Senate confirms Janet Yellen as chair of US Federal Reserve} \textbf{sentence 2}: \textit{US Senate Confirms Janet Yellen as New Central Bank Chief} & 4.2 & No \\
        STSB & \textbf{sentence 1}: \textit{Senate confirms \textcolor{blue}{John} Yellen as chair of US Federal Reserve} \textbf{sentence 2}: \textit{US Senate Confirms \textcolor{blue}{John} Yellen as New Central Bank Chief} & 4.2 & Yes \\
        \bottomrule
  \end{tabular}
 % \end{adjustbox}
  \end{center}
  \caption{Original and perturbed examples from the GLUE tasks.}
  \label{tab:glue-examples}
\end{table*}

\section{Preserving Classification Labels After Perturbation}\label{sec:preservelabel}

We have assumed for the purposes of the \fairscore~that perturbing word axes and attributes should not affect the gold classification label. In general, this is a reasonable assumption, but there are edge cases, in particular, for examples that rely on human-denoting references as part of their meaning. Consider for example the hypothetical textual entailment example \{P: \textit{John saw his aunt}, H: \textit{John saw his uncle}, gold-label: \texttt{not-entailment}\}. %For RTE/NLI in particular, we concatenated the premise and the hypothesis so that annotators would perturb all relevant references to the same entity together---if \textit{John} were the chosen word and the target attribute was \texttt{race/ethn:african-american}, we might get \{P': \textit{Jamal saw his aunt}, H: \textit{Jamal saw his uncle}, gold-label: \texttt{not-entailment}\}, preserving the original gold label through the perturbation. However, 
If \textit{aunt} is the chosen word, and the target attribute is \texttt{gender:man}, we have an issue: the new example will be \{P: \textit{John saw his uncle}, H: \textit{John saw his uncle}, gold-label: \textbf{\texttt{entailment}}\}. The entailment label will have changed, because the original example relied on the contrast of \textit{aunt} and \textit{uncle}, and even though we concatenated the premise and the hypothesis so coreference across them would be clear, the perturbation still changed the gold label in this hypothetical example. 

To get an estimate of how much perturbation actually altered the ground truth classification for our investigated tasks, we ran a pilot hand-validation of a subset of perturber perturbed examples from RTE, CoLA, SST-2, QNLI, QQP.\footnote{STS-B was excluded because it is on a 5 point Likert scale that was averaged over several annotators such that many examples have fractional scores. We found it hard with only a single pilot annotation to determine how close was close enough to count as gold label agreement.} We enlisted one expert annotator and instructed them to label, or validate 25 randomly selected perturbed examples per task, for a total of 125 examples. See \autoref{tab:glue-examples} for examples. The validator labels agreed with the original gold labels for the majority of the examples: 25/25 RTE examples, 25/25 CoLA examples, 25/25 SST-2 examples, 21/25 QNLI examples, and 20/25 QQP examples. 

Generally, when the validator label didn't agree with the gold, there was noise in the source data. For example, in QNLI, \textit{In which year did Alexander Dyce bequeathed his books to the museum?} was listed as entailing \textit{These were bequeathed with over 18,000 books to the museum in 1876 by John Forster.}, although the bequeather of the books differs across the two sentences in the source (the perturber only changed ``John'' to ``Jay''). QQP was somewhat of an outlier in our pilot validation, because it has a unexpectedly high proportion of explicit sexual content, which resulted in more drastic semantic changes for the 5 examples the validator disagreed on. 

In short, the methodological assumption that demographic perturbation shouldn't alter the gold label seems largely warranted, although we might take the QQP results with a grain of salt. A more in-depth validation round could be performed to confirm our pilot findings. %The expert validator was only provided with the premise-hypothesis pairs, not the original label, and they provided a label from \{entailment, contradiction, neutral\}. 

\end{document}